\documentclass[lettersize,journal]{IEEEtran}
\usepackage{amsmath,amssymb,amsfonts}
\usepackage{algorithmic}
\usepackage{algorithm}
\usepackage{array}
\usepackage[caption=false,font=footnotesize,labelfont=rm,textfont=rm]{subfig}
\usepackage{textcomp}
\usepackage{stfloats}
\usepackage{url}
\usepackage{verbatim}
\usepackage{graphicx}
\usepackage{cite}

\usepackage{color}
\usepackage{bm}
\usepackage{makecell}
\usepackage{enumerate}
\usepackage{amsthm}
\usepackage{mathrsfs}
\usepackage{xcolor}%
\usepackage{textcomp}%
\usepackage{manyfoot}%
\usepackage{listings}
\usepackage{booktabs}
\RequirePackage{multirow}
\RequirePackage{float}
\RequirePackage[numbers]{natbib}
\begin{document}

\title{Learning nonparametric DAGs with incremental information via high-order HSIC}

\author{Yafei WANG, Jianguo LIU
\thanks{$\bullet$ Yafei WANG is with School of Information Management and Engineering, Shanghai University of Finance and Economics, Shanghai, P.R.China. E-mail: yafei\_wang@163.sufe.edu.cn}
\thanks{$\bullet$ Jianguo LIU is corresponding author and with Institute of Accounting and Finance, Shanghai University of Finance and Economics, Shanghai, P.R.China. He is also with Research Group of Computational and AI Communication at Institute for Global Communications and Integrated Media, Fudan University, Shanghai, P.R.China. E-mail: liu.jianguo@sufe.edu.cn.}}

\markboth{Journal of \LaTeX\ Class Files,~Vol.~14, No.~8, August~2021}%
{Shell \MakeLowercase{\textit{et al.}}: A Sample Article Using IEEEtran.cls for IEEE Journals}


\maketitle

\begin{abstract}
Score-based methods for learning Bayesain networks(BN) aim to maximizing the global score functions. However, if local variables have direct and indirect dependence simultaneously, the global optimization on score functions misses edges between variables with indirect dependent relationship, of which scores are smaller than those with direct dependent relationship. In this paper, we present an identifiability condition based on a determined subset of parents to identify the underlying DAG. By the identifiability condition, we develop a two-phase algorithm namely optimal-tuning (OT) algorithm to locally amend the global optimization. In the optimal phase, an optimization problem based on first-order Hilbert-Schmidt independence criterion (HSIC) gives an estimated skeleton as the initial determined parents subset. In the tuning phase, the skeleton is locally tuned by deletion, addition and DAG-formalization strategies using the theoretically proved incremental properties of high-order HSIC. Numerical experiments for different synthetic datasets and real-world datasets show that the OT algorithm outperforms existing methods. Especially in Sigmoid Mix model with the size of the graph being ${\rm\bf d=40}$, the structure intervention distance (SID) of the OT algorithm is 329.7 smaller than the one obtained by CAM, which indicates that the graph estimated by the OT algorithm misses fewer edges compared with CAM. Source code of the OT algorithm is available at https://github.com/YafeiannWang/optimal-tune-algorithm.
\end{abstract}

\section{Introduction}
\IEEEPARstart{L}{earning} a DAG or a Bayesian network from data is to capture a graph that represents a probabilistic knowledge on dependencies over a set of variables ${\rm\bf X}=\left\{{\rm X}_1,...,{\rm X}_d\right\}$, where each node corresponds to a variable with an associated state distribution and each arc expresses dependence relationship \citep{Jensen2001}. Understanding causal relationships and effects is one of the fundamental goals of science \cite{Vowels2021}, with a diverse set of applications in risk assessment  \citep{Fenton2012, Dong2021, Yang}, finance \citep{Sanford2008, Liu2022}, recommendation system \citep{Zheng2021}, as well as feature selection \citep{Yu2020}. Traditionally, the gold standard to discover causal relationships remains interventions or randomized experiments, while it is often time-consuming or even infeasible in practice \cite{Vowels2021}. To deal with infeasibility in practice, a growing attention pays to inferring dependent relations from passively observational data, which is a crucial and classical issue in machine learning referring to learning DAG.

It is difficult to learn a DAG from data directly since it is an NP-hard problem \citep{Chickering2004}. A number of publications have tackled the problem to search the optimal structure according to score functions, which can be sub-categorized to combinatoric graph search and continuous optimization. Combinatoric graph search aims to discovering the optimal structure according to some scoring function. GES searches heuristically the structure which minimizes the BDeu score on the data \citep{Chickering2003}. Causal additive model (CAM) learns a DAG through sparse additive regression based on the estimated order search \citep{Buhlmann2014}. CAM simplifies the structure search by preliminary neighborhood selection and restriction on the structure of the superset of the skeleton using maximum likelihood estimation. However, the combinatoric graph search, such as GES and CAM algorithms, needs to make assumptions on a proper model class. Gao et al. provide a nonparametric identifiability based on variance between variables and learns the DAG by minimizing the residual variances, which simplifies the combinatoric graph search\citep{Gao2020}. Ye et al. propose a regularized Cholesky score to learn the DAG, which searches permutations of nodes by the sparse Cholesky factorization problem and outperforms in large graphs \citep{Ye2021}.
\begin{figure*}[t]
	\centering
	\includegraphics[scale=0.34]{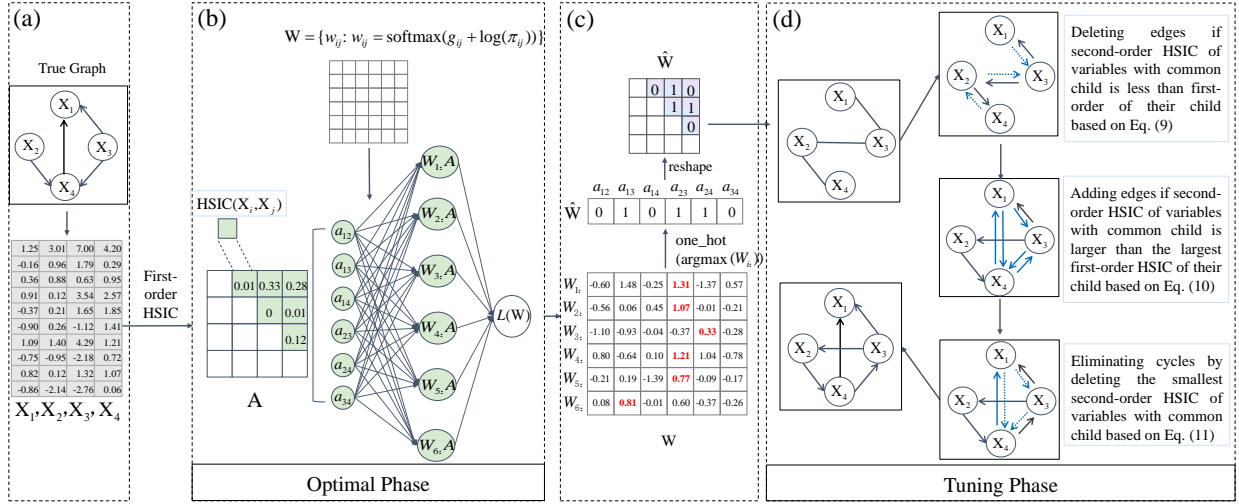}
	\caption{(Color online) The framework of the optimal-tuning (OT) algorithm. The subplot (a) shows the true graph with $d=4$ variables, where only the generated variables without structure information could be observed directly. In the first phase in subplot (b), namely optimal phase, we exploit the optimal combination of $\rm\bf{X}$ (generated according to the true graph shown in subplot (a)) based on first-order HSIC matrix $\rm\bf{A}$ to reach a global optimization. To transform discrete problem on first-order HSIC selection to a continuous optimization, we use Gumbel-softmax trick to generate parameter masked matrix $\rm\bf{W}$, where each row of matrix $W_{i:}$ is a selection vector to pick one of first-order HSIC between variables with the most largest elements in the row. Each elements of matrix $w_{ij}$ is denoted by a differentiable function $\text{softmax}(g_{ij}+\text{log}(\pi_{ij}))$, where $g_{ij}$ is drawn from Gumbel distribution with location 0 and scale 1, $\pi_{ij}$ is produced by a softmax function $\text{exp}(z_{ij})/\sum _{j}\text{exp}(z_{ij})$ and $z_{ij}$ is unconstrained parameters to be trained. By minimizing $L({\rm\bf W})$, as is shown in subplot (c), we obtain the optimal skeleton $\hat{\rm\bf W}$ using the one hot function for each row $W_{i:}$. In subplot (c), the matrix elements of $\hat{\rm\bf W}$ is 1 if the corresponding element $a_{ij}$ shown in subplot (b) is selected, which are calculated using the one hot function for each row $W_{i:}$. In the second phase in subplot (d), namely tuning phase, we tune the skeleton with deletion, addition and DAG-formalization strategies based on the theoretical proved incremental properties of first-order and second-order HSIC in Proposition 4.}
	\label{fig:2}
\end{figure*}

By applying a nonparametric estimators to a continuous score function for evaluating the structural equations, continuous optimization methods translate the combinatoric graph search to a continuous optimization problem. For example, a line of work embeds neural networks into dependent search algorithms based on a continuous DAG constraint. NOTEARS \citep{Zheng2018} is the first to propose a well-defined function for characterizing acyclicity and solve the least-squares program with an $L1$ penalty to lead to sparsity on linear models. NOTEARS+ extends NOTEARS acyclicity constraint to nonlinear model using partial derivatives of neural network as an independent criteria to formulate dependency structure \citep{Zheng2020}. In addition to using log-likelihood and least square score as an optimal goal, continuous independent criteria are the other way to learn DAG with nonparametric estimators on general additive models. Zhu and Chen minimize the BIC between variables as an optimal goal with DAG constraint \citep{Zhu2019}.

However, the score-based metheds focus on the global optimization regardless of indirect dependence where the local variables may have direct and indirect dependence simultaneously, leading to missing edges. We give a 4-variable example of Casaul additive models (CAM)\citep{Buhlmann2014}. The reason we present the CAM algorithm as an example lies in the fact that it applies in nonparametric settings which obtain a global minimizer without theoretical guarantee. The CAM algorithm searches the underlying DAG by greedily adding edges according to the score function $\sum_{j=1}^{d}{\rm log}({\rm E}({\rm Var}({\rm X}_j|PA({\rm X}_j))))$. For example, initially, CAM searches for the edge $({\rm X}_i,{\rm X}_j)$ by maximizing the function ${\rm log}({\rm E }({\rm Var}({\rm X}_i)))-{\rm log}({\rm E }({\rm Var}({\rm X}_i|{\rm X}_j)))$, which corresponds to find the maximal log-differences ${\rm log}({\rm E }({\rm Var}({\rm X}_i|{\rm X}_S))))-{\rm log}({\rm E }({\rm Var}({\rm X}_i|{\rm X}_{S\cup j})))$ at each iteration until the estimated graph is determined. However, if one pair of variables has the direct and indirect dependence simultaneously, the conditional variance of indirect dependent variable would be smaller than the one of the direct dependent variable leading to missing edges based on the global minimizer. Consider a graph with four edges $\{{\rm X}_{2}\rightarrow {\rm X}_{4}, {\rm X}_{3}\rightarrow {\rm X}_{1}, {\rm X}_{3}\rightarrow {\rm X}_{4}, {\rm X}_{4}\rightarrow {\rm X}_{1}\}$ based on additive noise model(ANM), which is shown in Fig. \ref{fig:2}(a). In this scenario, the dependent variables for ${\rm X}_{1}$ are ${\rm X}_{3}$ and ${\rm X}_{4}$. Moreover, variable ${\rm X}_{1}$ has indirect dependent relationships with ${\rm X}_{2}$ and ${\rm X}_{3}$ in items of the common child ${\rm X}_{4}$, which suggests that variable ${\rm X}_{4}$ is also dependent on ${\rm X}_{2}$ and ${\rm X}_{3}$. When ${\rm X}_{1}$ is selected at first, the edge ${\rm X}_{3}\rightarrow {\rm X}_{1}$ is not selected in the second step as the score function ${\rm log}({\rm E }({\rm Var}({\rm X}_1))-{\rm log}({\rm E }({\rm Var}({\rm X}_1|{\rm X}_4)))$ is larger than the one ${\rm log}({\rm E }({\rm Var}({\rm X}_1)))-{\rm log}({\rm E }({\rm Var}({\rm X}_1|{\rm X}_3)))$. The reason is that when ${\rm X}_4$ is dependent on ${\rm X}_3$, the variance of ${\rm X}_3,{\rm X}_4$ based on ANM ${\rm Var}({\rm X}_3+{\rm X}_4)$ is larger than the one ${\rm Var}({\rm X}_3)$ according to condition variance identity\citep{casella2021}. In the third step, the edge ${\rm X}_{3}\rightarrow {\rm X}_{1}$ is also not selected as the score function ${\rm log}({\rm E }({\rm Var}({\rm X}_1|{\rm X}_4))))-{\rm log}({\rm E }({\rm Var}({\rm X}_1|{\rm X}_{4\cup 2})))$ is larger than the one ${\rm log}({\rm E }({\rm Var}({\rm X}_1|{\rm X}_4))))-{\rm log}({\rm E }({\rm Var}({\rm X}_1|{\rm X}_{4\cup 3})))$. The reason is that when ${\rm X}_4$ is dependent on ${\rm X}_3$, the variance of ${\rm X}_3,{\rm X}_4$ based on ANM ${\rm Var}({\rm X}_2+{\rm X}_3+{\rm X}_4)$ is larger than the one ${\rm Var}({\rm X}_2+{\rm X}_3)$ according to condition variance identity. Finally, CAM returns the estimated graph with a missing edge ${\rm X}_{3}\rightarrow {\rm X}_{1}$. Therefore, to distinguish the direct dependence from indirect dependence, it is necessary to generate a framework for tuning the optimal solution that does not violate the constraint on DAG structure.

In this paper, we present an identifiability condition based on a determined parents subset for ANMs to identify the underlying DAG by comparing different order of dependent level among variables. Based on the identifiability condition, we propose an optimal-tuning algorithm, namely OT, using theoretically proved incremental properties of high-order HSIC, which captures the edges between variables with indirect dependence by locally amending the global optimization. A 4-variable example of the OT algorithm is given in Fig. \ref{fig:2}. Consider the graph as is shown in Fig. \ref{fig:2}(a), there are $4!=24$ possible situations ${\rm for}\ d=4$ and all of them together represent a full-connected graph. If we know a determined subset of parents of each variables, such as the determined subset of parent $PA_d({\rm X_1})=\{{\rm X_4}\}$ for ${\rm X_1}$ and $PA_d({\rm X_4})=\{{\rm X_2}\}$ for ${\rm X_4}$, then one can identify the underlying DAG by comparing the different order of dependence level, which is theoretically proved by Theorem 1. Unlike existing score-based methods that the process of optimal structure learning only account for the score of direct dependence, the proposed OT learns an optimal skeleton as the initial determined parents subset in the optimal phase and amend the skeleton with local deletion, addition and DAG-formalization strategies in the tuning phase. More concretely, we delete the redundant edges based on a smaller second-order HSIC value comparing with first-order one, add missing edges based on a larger second-order HSIC value comparing with first-order one and constraint the output being a DAG based on a larger second-order HSIC between each pair of variables. Consider the graph as is shown in Fig. \ref{fig:2}(a), the OT algorithm can identify the edge ${\rm X}_{3}\rightarrow {\rm X}_{1}$ in terms of the local addition strategy based on the incremental information of high-order HSIC among variables. The proposed algorithm provides an incremental dependence criterion for learning nonparametric DAGs with statistical guarantees, which can reduce missing edges for learning DAG.

\section{Preliminaries}
\subsection{Model definition}\label{subsec2}

We assume that the generation process of observed data is based on the structure equation model described below \citep{Judea2003}. Let ${\rm\bf X}=\{{\rm X}_1,...,{\rm X}_d\}$ denote a random vector with $d$ dimension taking value in measureable space $\Omega_{{\rm X}_i}$ and $\mathcal{G}=(V,E)$ a DAG where each random variable ${\rm X}_i$ is associated with a vertex $i$  in $V$. $PA({\rm X}_i)$ denotes the parent set of a variable ${\rm X}_i$ where $PA({\rm X}_i)=\left\{j:(j,i)\in E\right\}$. To formulate the dependent mechanisms between variables as structure equation model, we assume that there exists function set $f=\left\{f_1,...,f_d\right\}$ for each $f_i: R^d\rightarrow R$ such that
\begin{equation}
{\rm X}_i=f_i(PA({\rm X}_i),e_{i}), \ for\ i=1,...,d,\label{eq:1}
\end{equation}
where $e_{i}$ is a random noise. Formally, the independence statement between variables means that for any ${\rm X}_k\notin PA({\rm X}_i)$, $e_{i}$ is constant for all random variables ${\rm\bf X}$. Our method is based on additive noise model (ANM) \citep{Peters2014}, which is a specific case when all the noise $e_{i}$ are additive. Further, we assume that$f_i(PA({\rm X}_i),e_{i})= f_{i}(PA({\rm X}_i))+e_{i}$ for $f_{i}(\cdot)$ being square integrable function in ANM. 

For the graph $\mathcal{G}$ called Bayesian network, it satisfies Markov condition for $\rm\bf{X}$, that each variable is conditionally independent of its non-effects variables given its direct parents. From the perspective on graph, a Bayesian network for $\rm\bf{X}$ can be indicated as an interpretation of the direct and indirect dependent relationship between variables, such as that an edge ${\rm X}_i\rightarrow {\rm X}_j$ means that ${\rm X}_i$ directly causes state of ${\rm X}_j$, and not vice versa. 

\subsection{Partial identifiability on nonparametric setting}
A general strategy in most literature to enforce identifiability is to set some assumptions on the structure of conditional distributions \citep{Peters2014} or compare residual variances through assuming or estimating dependence mechanism \citep{Gao2020}. Our first result indicates that identifiability is guaranteed if a subset of parents of each variable is determined (without any structural assumptions or estimations) and only need to compare different order of dependence level among variables. First,we need to define the dependence level between variables and the order of dependence level.

The covariance of variables ${\rm X_{i}}$ and ${\rm X_{j}}$ is a statistical test of independence if the relationship between ${\rm X_{i}}$ and ${\rm X_{j}}$ is linear. However, when the situation is extend to nonlinear relationship, we need to emunerate different mapping of variables ${\rm X_{i}}$ and ${\rm X_{j}}$ to measure the dependence level of variables. More formally, we give the definition as follows:
 
\noindent
{\bf Definition 1} {\it Let $\mathcal{H}_{{ X}_1}$ be the space of measurable real functions $h$ defined on $\Omega_{{ X}_1}$ and $\mathcal{H}_{{ X}_{-1}}$ be the space analogous to $\mathcal{H}_{{ X}_1}$ with respect to ${\bf X}_{-1}=(X_2,...,X_d)$, then the dependence level between $X_1$ and ${\bf X}_{-1}$ is defined as:}\\
\begin{equation}
\begin{aligned}
&dep({\rm X_1}, {\rm\bf X}_{-1})=\\
&\sum_{h\in \mathcal{H}_{{\rm X}_1} ,h'\in \mathcal{H}_{{\rm\bf X}_{-1}}}\left|{\rm E}(h({\rm X_1})h'({\rm\bf X}_{-1}))-{\rm E}h({\rm X_1}){\rm E}h'({\rm\bf X}_{-1})\right|.
\end{aligned}
\end{equation}

where functions $h,h'$ are the mapping of variables ${\rm X_{1}}$ and ${\rm\bf X_{-1}}$ from space $\Omega_{\rm X_{1}}$ and $\Omega_{\rm\bf X_{-1}}$ to $R$ respectively. Note that functions $h,h'$ is used to capture the nonlinear relationship between variables represented by functions $f$ by sampling functions in a function space $\mathcal{H}_{{ X}_i}$. The definition of dependence level of variables ${\rm X_{1}}$ and ${\rm\bf X_{-1}}$ is an extension of the covariance of ${\rm X_{1}}$ and ${\rm\bf X_{-1}}$ to a function space. The covariance of variables ${\rm X_{1}}$ and ${\rm\bf X_{-1}}$ measures the independence only for linear relationship between variables ${\rm X_{1}}$ and ${\rm\bf X_{-1}}$, which can not measure the independence under nonlinear relationship between each pair of variables. Therefore, $dep({\rm X_1}, {\rm\bf X}_{-1})$ is the sum of the covariance of different functions $h({\rm X_1})$ and $h'({\rm\bf X}_{-1})$, which can capture the dependence level for nonlinear relationship. If variables ${\rm X_{1}}$ and ${\rm\bf X_{-1}}$ are independent, the expectation of joint distribution $({\rm X_{1}},{\rm\bf X_{-1}})$ for any functions $h({\rm X_1})$ and $h'({\rm\bf X}_{-1})$ equals the expectation of margin distribution ${\rm X_{1}}$ times the expectation of margin distribution ${\rm\bf X_{-1}}$ for any functions $h({\rm X_1})$ and $h'({\rm\bf X}_{-1})$. That is, the $dep({\rm X_1}, {\rm\bf X}_{-1})=\sum_{h,h'}\left|{\rm E}(h({\rm X_1})h'({\rm\bf X}_{-1}))-{\rm E}h({\rm X_1}){\rm E}h'({\rm\bf X}_{-1})\right|$ equals zero when ${\rm\bf X_{1}}$ and ${\rm\bf X_{-1}}$ are independent. When ${\rm X_{1}}$ and ${\rm\bf X_{-1}}$ are dependent, the dependence level $dep({\rm X_1}, {\rm\bf X}_{-1})$ is larger than zero due to the difference between the expectation of joint distribution $({\rm X_{1}},{\rm\bf X_{-1}})$ and the expectation of margin distribution ${\rm X_{1}}$ times the expectation of margin distribution ${\rm\bf X_{-1}}$ for functions $h({\rm X_1})$ and $h'({\rm\bf X}_{-1})$. Moreover, the order of dependence level $dep({\rm X_1}, {\rm\bf X}_{-1})$ is defined as the number of variables in the variables set ${\rm\bf X}_{-1}$. For example, the order of dependence level for $dep({\rm X_{1}},({\rm X_{i}},{\rm X_{j}}))$ is two.

\begin{figure}[t]
	\centering
	\includegraphics[scale=0.6]{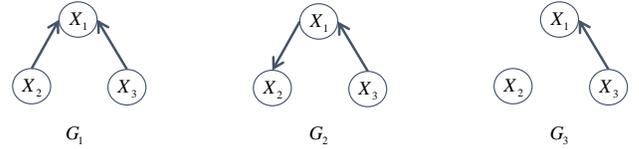}
	\caption{$G_1$: ${\rm X}_{1} = {\rm X}_{2}+{\rm X}_{3}+e_1, {\rm X}_{2} = e_2$ and ${\rm X}_{3}=e_3$; $G_2$: ${\rm X}_{1} = {\rm X}_{3}+e_1, {\rm X}_{2} = {\rm X}_{1}+e_2$ and ${\rm X}_{3}=e_3$; $G_3$: ${\rm X}_{1} = {\rm X}_{3}+e_1, {\rm X}_{2} = e_2$ and ${\rm X}_{3}=e_3$, where $e_i \sim N(\epsilon_i,\sigma_i^2)$ for all $i\in\{1,2,3\}$.}
	\label{fig:6}
\end{figure}

Furthermore, in terms of Definition 1, we prove Theorem 1, which is an identifiability condition. As an example, a graph with three variables ${\rm\bf X}=\{{\rm X_1,X_2,X_3}\}$ is given in Fig. \ref{fig:6}, in which variable ${\rm X}_3$ is the parent of variable ${\rm X}_1$ for all graphs ${\rm G}_i(i=1,2,3)$ and functions $f$ are linear. However, there are three type relations between variables ${\rm X}_1$ and ${\rm X}_2$.

The key point to determine whether $G_1$, $G_2$ or $G_3$ is the underlying graph is to distinguish their second-order dependence level $dep({\rm X_1},({\rm X}_{2},{\rm X}_{3}))$ from first-order dependence level $dep({\rm X_1},{\rm X}_{3})$. For $G_1$, if ${\rm X}_{3}$ is determined as one of ${\rm X_1}$'s parents, we can see the following results from the condition variance identity:
\begin{enumerate}
\item $dep({\rm X_1},{\rm X}_{3}) = \sum_{h \in \mathcal{H}_{{\rm X}_1},h'\in \mathcal{H}_{{\rm X}_3}}|E(h({\rm X_1})h'({\rm X}_{3}))-Eh({\rm X_1})Eh'({\rm X}_{3})|$\\$ \leq \sum_{h''\in  \mathcal{H}_{({\rm X}_2, {\rm X}_3)}}{\rm Var}(h''({\rm X}_{2},{\rm X}_{3}))$, and
\item $dep({\rm X_1},({\rm X}_{2},{\rm X}_{3})) = $ \\ $ \sum_{h\in \mathcal{H}_{{\rm X}_1},h''\in  \mathcal{H}_{({\rm X}_2, {\rm X}_3)}}|E(h({\rm X_1})h''({\rm X}_{2},{\rm X}_{3}))- Eh({\rm X_1})Eh''({\rm X}_{2},{\rm X}_{3})|$\\$= \sum_{h''\in  \mathcal{H}_{({\rm X}_2, {\rm X}_3)}}{\rm Var}(h''({\rm X}_{2},{\rm X}_{3}))$.
\end{enumerate}

The above result 1 indicates that the dependence level between variables ${\rm X_1}$ and ${\rm X}_{3}$ is smaller than the dependence level between $({\rm X}_{2},{\rm X}_{3})$ and themselves. Based on the covariance inequality, $\left|E(h({\rm X_1})h'({\rm X}_{3}))-Eh({\rm X_1})Eh'({\rm X}_{3})\right|$ is smaller than the square root of ${\rm Var}(h'({\rm X}_{3})){\rm Var}(h''({\rm X}_{2},{\rm X}_{3}))$. According to the condition variance identity\citep{casella2021}, we can unfold ${\rm Var}(h'({\rm X}_{3}))$ to ${\rm E}({\rm Var}(h'({\rm X}_{3})|{\rm X}_{2},{\rm X}_{3}))+{\rm Var}({\rm E}(h'({\rm X}_{3})|{\rm X}_{2},{\rm X}_{3}))$. Given $({\rm X}_{2},{\rm X}_{3})$, the expectation of condition variance ${\rm Var}(h'({\rm X}_{3})|{\rm X}_{2},{\rm X}_{3})$ is equal to ${\rm E}((h'({\rm X}_{3})-{\rm E}(h'({\rm X}_{3})|{\rm X}_{2},{\rm X}_{3}))^2)=0$ due to $h'({\rm X}_{3})$ can be represented by $h''({\rm X}_{2},{\rm X}_{3})$ for the space $\mathcal{H}_{{\rm X}_3} \subset \mathcal{H}_{({\rm X}_2, {\rm X}_3)}$\citep{bobrowski2005} and the variance of condition expectation ${\rm E}(h'({\rm X}_{3})|{\rm X}_{2},{\rm X}_{3})$ is equal to ${\rm Var}(h''({\rm X}_{2},{\rm X}_{3}))$. One can obtain the result 1 intuitively because ${\rm X_1}$ is represented by both ${\rm X_2}$ and ${\rm X}_{3}$ according to the ANM defined in Section II, which gives rise to a smaller $dep({\rm X_1},{\rm X}_{3})$ compared to the dependence level between $({\rm X}_{2},{\rm X}_{3})$ and themselves. Similarly, the result 2 indicates that the dependence level between ${\rm X_1}$ and $({\rm X}_{2},{\rm X}_{3})$ is equal to the dependence level between $({\rm X}_{2},{\rm X}_{3})$ and themselves. It can be obtained intuitively because ${\rm X_1}$ can be represented  by $({\rm X}_{2},{\rm X}_{3})$ according to the structure equation model defined in Section II. Hence, we can infer that $dep({\rm X_1},{\rm X}_{3})\leq dep({\rm X_1},({\rm X}_{2},{\rm X}_{3}))$ for the graph $G_1$ as long as one of ${\rm X_1}$'s parents ${\rm X}_{3}$ is determined.

In the same manner, we can infer that $dep({\rm X_1},{\rm X}_{3})\geq dep({\rm X_1},({\rm X}_{2},{\rm X}_{3}))$ for $G_2$. Lastly, for the graph $G_3$, the dependence level between variables ${\rm X_1}$ and ${\rm X}_{3}$ is equal to the dependence level between ${\rm X}_{3}$ and itself, that is,  $dep({\rm X_1},{\rm X}_{3})=\sum_{h'}{\rm Var}(h'({\rm X}_{3}))$. We can approximate $h''({\rm X}_{2},{\rm X}_{3})$ as $h({\rm X}_{2})+h'({\rm X}_{3})+h({\rm X}_{2})h'({\rm X}_{3})$ by Taylor polynomial of a function of two variables. Then, the dependence level between ${\rm X_1}$ and $({\rm X}_{2},{\rm X}_{3})$ is ${\rm Var}(h'({\rm X}_{3}))+{\rm E}h({\rm X}_{2}){\rm Var}(h'({\rm X}_{3}))$. There is no guarantee that the value of ${\rm E}h({\rm X}_{2})$ is larger than zero or not. Hence, the $dep({\rm X_1},{\rm X}_{3})$ can be larger or smaller than $dep({\rm X_1},({\rm X}_{2},{\rm X}_{3}))$ for $G_3$.

The existing of an edge of the graph could be calculated in terms of the dependence level between variables ${\rm X_1}$ and ${\rm X}_{2}$. For the graph $G_3$, $dep({\rm X_1},{\rm X}_{2})$ is zero due to the fact that variables ${\rm X_1}$ and ${\rm X}_{2}$ are independent, whereas for graphs $G_1$ and $G_2$, $dep({\rm X_1},{\rm X}_{2})$ is larger than zero. Therefore, we can identify the underlying graph by comparing the first-order dependence level $dep({\rm X_1},{\rm X}_{3})$ and the second-order dependence level $dep({\rm X_1},({\rm X}_{2},{\rm X}_{3}))$.

From the 3-variable example, the conclusion could be extended to the $d$-variable ANMs. Since both the covariance inequality and condition variance identity do not require the assumptions of linearity of $f$ and Gaussian noise distributions, the main extension on model identifiability from the three variables to multivariate involves two parts: The determined subset of variables' parents and the comparisons between the different dependence level.

\noindent
{\bf Theorem 1} {\it Let joint probability $P({\rm\bf X})$ be generated from a DAG $G$ with ANM \eqref{eq:1} and $PA_d({\rm X}_i))$ be the determined parents of ${\rm X}_i$. If random noise $e_{i}$ is independent on variable ${\rm X}_i)$'s parents and the number of $PA_d({\rm X}_i))$ is one at least, then the Bayesian network is identifiable from joint probability $P(\rm\bf{X})$.} 

The detailed proof of Theorem 1 is provided in Appendix A. Theorem 1 proves that ANMs are identifiable when one of the parent variable information is determined. To a third variable ${\rm X}_d$, ${\rm X}_d$ is the parent for variable ${\rm X}_i$ when the dependence level $dep({\rm X}_i,({\rm X}_j,{\rm X}_d))$ is larger than the one $dep({\rm X}_i,{\rm X}_j)$, where ${\rm X}_j$ is the determined parent for ${\rm X}_i$. That is, the dependence level between ${\rm X}_i$ and its parents is the largest compared with the dependence level between ${\rm X}_i$ and subset of its parents. Thus, we can identify the underlying graph by searching the most dependent component for each variables based on the incremental change between different order of dependence level among variables.

According to Theorem 1, for the presented OT algorithm, there are two essential points for identifiability of ANMs: 
\begin{enumerate}
\item Dependence level. How to calculate the different order of dependence level in a function space when the form of function $h(\cdot)$ is unknown.
\item Determined parent set. How to get and update the initial determined parents set for each variables.
\end{enumerate} 
The first point requires a cross-covariance operator on a space of functions from ${\rm\bf X}$ to space ${\bf R}$. Therefore, in Section III, we provide the Hilbert-Schmidt independence criterion (HSIC) on additive reproducing kernel Hilber space as the cross-covariance operator. The second point requires a heuristic strategy to get and update the initial determined parents set for each variables, according to the incremental change between different order of dependence level. Therefore, in Section III, we theoretically analyze the incremental properties between different order HSIC. And in Section IV, we provide a two-phase method namely OT using different order of HSIC theoretically proved in Section III. In the optimal phase, the OT algorithm initializes the determined parents of each variables. In the tuning phase, the OT algorithm updates the set of determined parents iteratively. 

\section{Hilbert-Schmidt independence criterion on additive noise model}
In this section, we provide the HSIC on additive reproducing kernel Hilbert space(RKHS) to measure the dependence level of variables defined in Section II, theoretically analyze the incremental properties between different order HSIC and postpone discussing implementation details to Section IV. To be more precisely understanding about the properties, we first review additive RKHS \citep{Lee2016} and cross-covariance on Hilbert space \citep{Gretton2005}. Then, we introduce the HSIC and extend it to an incremental measurement with respect to independent criterion on ANM, which is a new perspective for independence detection and will be used to learn DAGs in Section IV.

\subsection{Additive reproducing kernel Hilbert space}
We briefly summarize notions in RKHS and its extension on additive form. Let ${\rm\bf X}$ be a $d$-dimensional random vector taking value in $\Omega$. Then consider $\mathcal{F}({\rm X}_i)$ as the map from space $\Omega_{{\rm X}_i}$ to $R$, if for each element $a \in \Omega_{{\rm X}_i}$, the Dirac evaluation operator $\delta_{{\rm X}_i}$ is a bounded linear functional mapping $f \in \mathcal{F}$ to $f({\rm X}_i) \in R$, without loss of generality, RKHS on variable ${\rm X}_i$ is defined as follows.

\noindent
{\bf Definition 2} (Reproducing kernel Hilbert space) {\it Let $\mathcal{H}_{{\rm X}_i} \in \mathcal{F}({{\rm X}_i})$ be an RKHS if there is an element $\phi_i(a) \in \mathcal{F}$ and a unique positive definite kernel $k_i:{\rm X}_i \times {\rm X}_i \rightarrow R$ satisfying $\left \langle \phi_i(a),\phi_i(b) \right \rangle_\mathcal{F}=k_i(a,b)$ for all elements $a, b \in \Omega_{{\rm X}_i}$.}

We suppose $k_i$ to be the same for $i=1,..,d$ and denote these kernels by $k$ as a common symbol. It has be proven that a positive semi-definite kernel $k$ is associated with an RKHS, of which inner product can be calculated by kernel $k$\citep{Baker1973}. Therefore, one can construct and utilize the relevant RKHS by using a positive semi-definite kernel.
Here we employ the Gaussian kernel on space $\Omega_{{\rm X}_i}$, a currently used positive semi-definite kernel, which is defined as $k({\rm X}_i,{\rm X}_i')=\text{exp}(-\frac{{\left\|{\rm X}_i-{\rm X}_i'\right\|}_{R}^{2}}{2\sigma^2}) \text{\ for\ all } {\rm X}_i\in \Omega_{{\rm X}_i}$. For the specification of the bandwidth parameter $\sigma$, we use median heuristic according to input data\citep{Zhang2018}, which set to the median of $\|{\rm X}_i-{\rm X}_i'\|^{2}_{R}$.

To measure the degree of correlation between variables considering linear and nonlinear effect, we first should generalize a statistic that efficiently summarizes the norm in RKHS. Here, we give the definition of Hilbert-Schmidt norm as the indicator.

\noindent
{\bf Definition 3} {(Hilbert-Schmidt norm) {\it Let $C:\ \mathcal{H}_{{\rm X}_i} \rightarrow \mathcal{H}_{{\rm X}_j}$ be a linear operator, the Hilbert-Schmidt norm of C is defined as:}
\begin{equation}
\left \| C \right \|_{HS}:=\sum_{t,s}(\left \langle Cv_t,u_s \right \rangle_{\mathcal{H}_{{\rm X}_i}}^{2})^{\frac{1}{2}},\label{eq:2}
\end{equation}
where $v_t$ and $u_s$ are orthonormal bases of spaces $\mathcal{H}_{{\rm X}_i}$ and $\mathcal{H}_{{\rm X}_j}$ respectively. The Hilbert-Schmidt norm is an extension of the Frobenius norm on matrices.}

One of the strengths of analyzing complicated objects of the RKHS is that using Hilbert space structure could embed them in a concise form, especially evaluating inner products through the reproducing property to simplify computation in a RKHS. In this paper, we introduce the embedding tool to analyze independence between variables ignoring assumption on probability distribution. Finally, we give the definition of additive RKHS, which will be used to characterize the difference in independence of multi-variables with ANM in Section II. 

\noindent
{\bf Definition 4} (Additive reproducing Hilbert space) {\it Let $\mathcal{H}_{X}$ be the direct sum $\bigoplus_{i=1}^{d}\mathcal{H}_{{\rm X}_i}$, such that $\mathcal{H}_X=\{f_1+...+f_d: \allowbreak f_1\in \mathcal{H}_{{\rm X}_1},...,f_d\in \mathcal{H}_{{\rm X}_d} \}$, with inner product $\left \langle  f_1+...+f_d, g_1+...+g_d\right \rangle_{\mathcal{H}_X}=\left \langle f_1,g_1\right \rangle_{\mathcal{H}_{{\rm X}_1}}+...+\left \langle f_d,g_d\right \rangle_{\mathcal{H}_{{\rm X}_d}}$}.

Our construction of summing the separable space directly as follows \cite{Aronszajn1950} .

\subsection{The cross-covariance operator on additive reproducing kernel Hilbert space}

We apply cross-covariance operator on RKHS to construct a criterion function for independence detection. The cross-covariance operator expresses the covariance between functions in the RKHS as a bilinear functional, and contains all the information regarding the dependence of X and Y expressible by nonlinear functions in the RKHS. While cross-covariance operator is generally used for measuring the independence of random variables on Banach spaces, the principle is much easier for RKHS \citep{Gretton2005}. This subsection only summarizes the basic definition and properties of cross-covariance operator and its extension on additive RKHS.

The following assumption on the kernel $k$ guarantees that the space $\mathcal{H}_{{\rm X}_i}$ is a subspace of $L_2(P_{{\rm X}_i})$, the class of all square-integrable functions of ${\rm X}_i$. Note that $L_2(P_X)$ is no itself a RKHS, a subspace $\mathcal{H}$ in $L_2(P_X)$ is a RKHS if and only if $\mathcal{H}$ has a reproducing kernel, and a RKHS is a subspace of $L_2(P_X)$ if $Ek(X,X)$ is finite.

\noindent
{\bf Assumption 1} {\it The expectation of kernel function ${\rm E}k({\rm X}_i,{\rm X}_i)<\infty,\ i=1,...,d$.}

This is a mild condition that holds for many commonly used kernels such as the Gaussian radial basis function. For Assumption 1, we can show that, for any $i=1,...,d$ and $j=1,...d$, there exists an operator $C_{{\rm X}_i{\rm X}_j}$ defined as:

\noindent
{\bf Definition 5} (Cross-covariance operator)  {\it According to Ref. \cite{Gretton2005}, the cross-covariance operator $C_{{\rm X}_i{\rm X}_j}:\mathcal{H}_{{\rm X}_j}\rightarrow\mathcal{H}_{{\rm X}_i}$ is a linear operator such that for all $f_i \in \mathcal{H}_{{\rm X}_i}$ and $f_j \in \mathcal{H}_{{\rm X}_j}$,}
\begin{equation}
\begin{aligned}
\left \langle f_{i}, C_{{\rm X}_i{\rm X}_j}f_j\right \rangle = &{\rm\bf E}_{{\rm X}_i{\rm X}_j}([f_i({\rm X}_i)-{\rm\bf E}_{{\rm X}_i}(f_i({\rm X}_i))]\\
&[f_j({\rm X}_j)-{\rm\bf E}_{{\rm X}_j}(f_j({\rm X}_j))]).\label{eq:3}
\end{aligned}
\end{equation}

Then the cross-covariance operator itself can be written
\begin{equation}
C_{{\rm X}_i{\rm X}_j}:={\rm\bf E}_{{\rm X}_i{\rm X}_j}[(\phi({\rm X}_i)-\mu_{{\rm X}_i})\otimes (\psi({\rm X}_j)-\mu_{{\rm X}_j})],\label{eq:4}
\end{equation}
where $\mu_{{\rm X}_i}:=\rm\bf{E}_{{\rm X}_i}\phi ({\rm X}_i)$, $\mu_{{\rm X}_j}:=\rm\bf{E}_{{\rm X}_j}\phi ({\rm X}_j)$, and $\otimes$ is tensor product defined on Hilbert space as a bilinear operator to complete the matrix space on two separable Hilbert space \citep{Fukumizu2009}. Let $P_{{\rm X}_i,{\rm X}_j}$, $P_{{\rm X}_i}$, $P_{{\rm X}_j}$ be the distributions of $({\rm X}_i,{\rm X}_j)$, ${\rm X}_i$, ${\rm X}_j$, respectively. As shown by \cite{Gretton2005}, the square Hilbert-Schmidt norm of $C_{{\rm X}_i,{\rm X}_j}$ is identical to the dependence level between the reproducing kernel Hilbert space embeddings of ${\rm X}_i$ and ${\rm X}_j$, which we will discuss in Subsection III C.

We generalize some notions of cross-covariance on additive reproducing kernel Hilbert space. Let $C_{{\rm\bf XX}}:\ \mathcal{H}_{\rm\bf X} \rightarrow \mathcal{H}_{\rm\bf X}$ be the matrix of operators whose $(i,j)$ the element is the operator $C_{{\rm X}_i{\rm X}_j}$. That is, for any $f=f_1+...+f_d \in \mathcal{H}_{\rm\bf X}$,
\begin{equation}
C_{\rm\bf XX}f=\sum_{j=1}^{d}\sum_{i=1}^{d}C_{{\rm X}_i{\rm X}_j}f_i.\label{eq:5}
\end{equation}

This structure was also used in \cite{Bach2008} and \cite{Lee2016}. We define the notation $C_{\rm\bf XX}$ as the additive covariance operator of ${\rm\bf X}$.

\subsection{ Hilbert-Schmidt independence criterion on ANM}

Building on the additive covariance operator, we now introduce HSIC to measure the dependence level between variables and apply HSIC to ANM for analyzing the incremental properties between different order of dependence level among variables. One possibility to measure the dependence level when the form of distribution function is unknown is to construct the HSIC \citep{Gretton2005} due to its no user-defined regularisation and convergence in the large sample limit. We first lay out the definition of HSIC using the square of the Hilbert-Schmidt norm of the cross-covariance operator.

\noindent
{\bf Definition 6} (Hilbert-Schmidt independence criterion) {\it Given separable reproducing kernel Hilbert spaces $\mathcal{H}_{{\rm X}_i}$, $\mathcal{H}_{{\rm X}_j}$ and a joint measure $p_{{\rm X}_i{\rm X}_j}$, the Hilbert-Schmidt Independence Criterion is defined as:}
\begin{equation}
\mbox{HSIC}(p_{{\rm X}_i{\rm X}_j},\mathcal{H}_{{\rm X}_i},\mathcal{H}_{{\rm X}_j}):=\left\|C_{{\rm X}_i{\rm X}_j}\right\|_{HS}^{2}.\label{eq:6}
\end{equation}
According to Ref. \cite{Gretton2005}, the Hilbert-Schmidt norm of $C_{{\rm X}_i{\rm X}_j}$ exists when the various expectations over the kernels are bounded, which is true as long as the kernels $k$ and $l$ are bounded as Assumption 1 indicated. Then HSIC can be expressed in terms of kernels as:
\begin{equation}
\begin{aligned}
&\text{HSIC}(p_{{\rm X}_i{\rm X}_j},\mathcal{H}_{{\rm X}_i},\mathcal{H}_{{\rm X}_j})=\\
&E_{{\rm X}_i,{\rm X}_{i}^{'},{\rm X}_j,{\rm X}_{j}^{'}}[k({\rm X}_i,{\rm X}_{i}^{'})l({\rm X}_j,{\rm X}_{j}^{'})]\\
&+E_{{\rm X}_i,{\rm X}_{i}^{'}}[k({\rm X}_i,{\rm X}_{i}^{'})]E_{{\rm X}_j,{\rm X}_{j}^{'}}[l({\rm X}_j,{\rm X}_{j}^{'})]\\ &-2E_{{\rm X}_i,{\rm X}_j}[E_{{\rm X}_{i}^{'}}k({\rm X}_i,{\rm X}_{i}^{'})E_{{\rm X}_{j}^{'}}l({\rm X}_j,{\rm X}_{j}^{'})].
\end{aligned}
\end{equation}
The Eq.(9) could be used to measure the dependence level between ${\rm X}_i$ and ${\rm X}_j$ simply  by considering expectation over kernel functions $k$ and $l$ on the separable Hilbert spaces regarding to the joint and marginal distributions in ${\rm X}_j$ and ${\rm X}_j$ without the density estimation assumption.

As HSIC is an indicator to measure the dependence level between ${\rm X}_i$ and ${\rm X}_j$, we simplify the symbol as $\text{HSIC}({\rm X}_i,{\rm X}_j)$. To apply HSIC to measure the dependence level defined in Definition 1, it is essential to clarify how to measure the dependence level between variables when variables can be represented by an additive noise function based on ANM and how to use HSIC to measure different order of dependence level among variables. We give the following propositions to clarify the above two questions. 

\noindent
{\bf Proposition 1 }{\it  Let $f_i\in \mathcal{H}_{i}$ be square integrable function, then $\text{HSIC}(f_i({\rm X}_i),f_i({\rm X}_j))$ can be approximated by $\text{HSIC}({\rm X}_i,{\rm X}_j)$.}

The detailed proof of Proposition 1 is provided in Appendix A. Proposition 1 supposes that the dependence level between each pair of variables is close to the dependence level between variables with function transformation. Another question we need to clarify when applying HSIC to ANM is how to use HSIC to measure different order of dependence level. We first give the definition of order of HSIC as follows:

\noindent
{\bf Definition 7} {\it For a set of variables ${\rm\bf X}=\{{\rm X}_1,...,{\rm X}_n\}$, the order of $\mbox{HSIC}({\rm X}_i,{\rm\bf X})$ refers to the number of variables in ${\rm\bf X}$}.

Definition 7 indicates that the order of dependence level is defined as the number of variables compared with the single variable. For example, the first order of HSIC for ${\rm X}_i$ refers to $\mbox{HSIC}({\rm X}_i,{\rm X}_j)$ and the second order of HSIC for ${\rm X}_i$ refers to $\mbox{HSIC}({\rm X}_i,({\rm X}_j,{\rm X}_t))$. After defining the order of HSIC, we now give an approximate of $\mbox{HSIC}({\rm X}_i,{\rm\bf X})$ based on additive RHKS, which is used to analyze the properties of different order between $\mbox{HSIC}({\rm X}_i,{\rm\bf X})$ and $\mbox{HSIC}({\rm X}_i,{\rm\bf X}\cup{\rm X}_j)$.

\noindent
{\bf Proposition 2} {\it $\mbox{HSIC}({\rm X}_i,{\rm\bf X})$ can be approximated by $\text{HSIC}({\rm X}_i,\sum_{j=1}^{n}{\rm X}_j)$ for ${\rm\bf X}=\{{\rm X}_1,...,{\rm X}_n\}$.} 

The detailed proof of Proposition 2 is provided in Appendix A. As we get the approximate representation of $\text{HSIC}({\rm X}_i,{\rm X}_j)$, we now analyze the properties for same order of dependence level between $\mbox{HSIC}({\rm X}_i,{\rm\bf X})$ and $\mbox{HSIC}({\rm X}_i,{\rm\bf X}')$, as is shown in Proposition 3, and properties for different order of dependence level between $\mbox{HSIC}({\rm X}_i,{\rm\bf X})$ and $\mbox{HSIC}({\rm X}_i,{\rm\bf X}\cup{\rm X}_j)$, as is shown in Proposition 4. We compare the different first-order of dependence level between $\mbox{HSIC}({\rm X}_i,{\rm X}_j)$ and $\mbox{HSIC}({\rm X}_i,{\rm X}_k)$, where ${\rm X}_i$ is dependent on ${\rm X}_j$ but not on ${\rm X}_k$. Intuitively, for the same order of dependence level, the value of $\text{HSIC}({\rm X}_i, {\rm X}_j)$ is larger if there exists a function associating these two variables, compared to two random variables. Then, one have the following propositions.

\noindent
{\bf Proposition 3 }{\it  If variables ${\rm X}_i$ and ${\rm X}_j$ are dependent while variables ${\rm X}_i$ and ${\rm X}_k$ are independent, then $\text{HSIC}({\rm X}_j, {\rm X}_i)>\text{HSIC}({\rm X}_k,{\rm X}_i)$.}

The detailed proof of Proposition 3 is provided in Appendix A. If variables only have direct dependent relationship with others, Proposition 3 shows that we can find an optimal combination of ${\rm \bf X}$ based on first-order HSIC as the skeleton of the underlying DAG. However, the indirect relationship between variables leads to indistinction of identifiability when only consider first-order HSIC. Consider a graph with four edges $\{{\rm X}_{2}\rightarrow {\rm X}_{4}, {\rm X}_{3}\rightarrow {\rm X}_{1}, {\rm X}_{3}\rightarrow {\rm X}_{4}, {\rm X}_{4}\rightarrow {\rm X}_{1}\}$, which is shown in Fig. \ref{fig:2}(a). In this scenario, ${\rm X}_{2}$ has an indirect relationship with ${\rm X}_{3}$ through ${\rm X}_{1}$ according to the DAG skeleton. Then there is no theoretical guarantee to distinguish the difference between $\text{HSIC}({\rm X}_1, {\rm X}_2)$ and $\text{HSIC}({\rm X}_1, {\rm X}_3)$ due to ${\rm X}_{4}$ is directly associated with ${\rm X}_1$ and can be represented by ${\rm X}_2$. In Section II, we discuss that after giving the initial determined parents set for each variables, the underlying DAG can be identified by comparing different order of dependence level. That is, if we know that ${\rm X}_{4}$ is one of ${\rm X}_{1}$'s parents, we can infer that the value of $dep({\rm X}_1, ({\rm X}_2,{\rm X}_4))$ is similar to $dep({\rm X}_1, {\rm X}_4)$ while the value of $dep({\rm X}_1, ({\rm X}_3,{\rm X}_4))$ is larger than $dep({\rm X}_1, {\rm X}_4)$, which indicates that variable ${\rm X}_3$ is one of ${\rm X}_{1}$ 's parents while ${\rm X}_2$ is not. Then we consider an incremental change between first-order HSIC and second-order HSIC to distinguish the different dependence level among variables, as is shown in Proposition 4.

\noindent
{\bf Proposition 4 } {\it Assume that variables ${\rm X}_i=f_i(PA({\rm X}_i))+e_i$ for some $f_i \in \mathcal{H}_{{\rm X}_i}$, we can infer for ${\rm X}_j, {\rm X}_s \in PA({\rm X}_i)$ and ${\rm X}_k \notin PA({\rm X}_i)$:
\begin{enumerate}[a)]
\item $\text{HSIC}(({\rm X}_j,{\rm X}_k),{\rm X}_i) \leq \text{HSIC}({\rm X}_j,{\rm X}_i)$;
\item $\text{HSIC}(({\rm X}_j,{\rm X}_s), {\rm X}_i) > max\left\{ \text{HSIC}({\rm X}_j,{\rm X}_i),\text{HSIC}({\rm X}_s,{\rm X}_i)\right\}$;
\item $\text{HSIC}(({\rm X}_j,{\rm X}_s),{\rm X}_i) >\text{HSIC}(({\rm X}_j,{\rm X}_k),{\rm X}_i)$.
\end{enumerate}}

The detailed proof of Proposition 4 is provided in Appendix A. According to Proposition 4, the incremental information of second-order HSIC compared with first-order HSIC indicates that if variable ${\rm X}_k$ is not the child of ${\rm X}_i$, its second-order with one of ${\rm X}_i$'s children ${\rm X}_j$ is smaller than the largest of their first-order HSIC. On the other hand,  if variable ${\rm X}_j$ and ${\rm X}_s$ are both the children of ${\rm X}_i$, their second-order HSIC is larger than the largest one obtained by the first-order HSIC. Moreover, if variable ${\rm X}_j$ and ${\rm X}_s$ are both the children of ${\rm X}_i$ and ${\rm X}_k$ is not the child of ${\rm X}_i$, the second-order HSIC among ${\rm X}_j$, ${\rm X}_s$ and ${\rm X}_j$ is larger than the second-order HSIC among ${\rm X}_j$, ${\rm X}_k$ and ${\rm X}_j$.

\section{Model estimation}
In this section, we present our two-phase method, namely OT. According to Theorem 1, the identifiability of the underlying DAG requires a heuristic strategy to initial the determined parents for each variables and iteratively update the determined parents according to the incremental change between different order of dependence level. Based on the theoretical proved incrementally properties of first-order and second-order HSIC in Section III, we give a two-phase method to obtain an optimal skeleton of underlying DAG as the determined parents set in the first phase using Proposition 3 and tune the skeleton regarding the difference between direct and indirect dependence in the second phase using Proposition 4. A framework of the proposed algorithm is shown in Fig. \ref{fig:2}.

\subsection{The optimal phase}

As is mentioned in Proposition 3, the dependence level measured by first-order HSIC is larger on pair variables associated with each other than the variables independent from each other. Although the indirect relationship between variables leads to indistinction of identifiability when only consider first-order HSIC, based on Theorem 1, we can solve the combinatorial optimization issues of DAG using first-order HSIC to learn the initial determined parents set, which is the skeleton of underlying DAG and will adjust in the tuning phase according to the theoretical proved incremental properties of first-order and second-order HSIC on Proposition 4. 

Instead of separating structure search from independent estimation, our method in the optimal phase exploits the optimal combination of ${\rm\bf X}$ based on first-order HSIC. Formally, let ${\rm\bf A}\in R^{s} $ for $s=d\times (d-1)/2$ be a vector with element $a_{t}$ being the value of $\text{HSIC}({\rm X}_i,{\rm X}_j)$ for  $i,j = 1,...,d$. Then the combination optimization issues can be formulated as the following program:
\begin{equation}
\begin{aligned}
&\min_{{\rm\bf W}\in R^{s\times s}} L({\rm\bf W}) \\
& \text{s.t.}\  L(\rm\bf{W})=\left\|({\rm\bf 1}-{\rm\bf W}) {\rm\bf A}\right\|_2^2+\lambda\left\|{\rm\bf W}\right\|_2. \label{eq:7}
\end{aligned}
\end{equation}

In the formula described above,  $\rm\bf{W}$ is a matrix with element $w_{i,j}\in\left\{0,1\right\}$, which is a selection mask on first-order HSIC of $\rm\bf{X}$. Intuitively, the formulation aims to find an optimal combination of dependence of $\rm\bf{X}$ based on first-order HSIC. Considering the discretization of program, it would cause computational issues with the number of variables increasing, we thus use a stochastic neural network to solve the problem, in which the gradient estimator is Straight Through (ST) gradient estimator \citep{Bengio2013} combined with Gumbel-Softmax trick \citep{Jang2017}. More concretely, we use Gumbel-softmax trick to transform $\rm\bf{W}$ to a parameter masked matrix with $w_{ij}=\text{softmax}(g_{ij}+\text{log}(\pi_{ij}))$, where each row of matrix $\rm{W}_{i:}$ is a selection vector to pick one of first-order HSIC between variables with the most largest elements in the row. Thus, we can select $s$ pair of variables based on first-order HSIC at most. To transform discrete problem on first-order HSIC selection to a continuous optimization, each elements of matrix $w_{ij}$ is denoted by a differentiable function $\text{softmax}(g_{ij}+\text{log}(\pi_{ij}))$, where $g_{ij}$ is drawn from Gumbel distribution with location 0 and scale 1, $\pi_{ij}$ is produced by a softmax function $\text{exp}(z_{ij})/\sum _{j}\text{exp}(z_{ij})$ and $z_{ij}$ is unconstrained parameters to be trained.  Then problem (\ref{eq:7}) can be embedded by a stochastic neural network and solved by the stochastic optimization method Adam proposed by Kingma and Ba \cite{Kingma2015}. As proved by Kingma and Ba\cite{Kingma2015}, the Adam method is convergent. Therefore, the computation complexity of problem (10) is ${\rm O}(d^3)$ for $d$ denoting the number of variables , which is about matrix multiplication of $({\rm\bf 1}-{\rm\bf W}) {\rm\bf A}$ as the Adam method is convergent.

By minimizing $L({\rm\bf W})$, we obtain the optimal graph $\hat{\rm\bf W}$ using the one hot function for each row $W_{i:}$. More concretely, $\hat{\rm\bf W}$ is a vector and the elements are equal to one if the corresponding elements in ${\rm\bf A}$ are selected. We adjust $\hat{\rm\bf W}$ to meet our specific requirement in the tuning phase. There exists a few works using Gumbel-Softmax trick to search for optimal structure \citep{Brouillard2020} and treating variable selection as an optimal object based on direct dependent score functions. We use first-order HSIC to measure direct dependent level instead of direct dependent score functions and use Gumbel-Softmax trick to select pair of variables according to first-order HSIC. In the next, we adjust the optimal structure in the tuning phase by the theoretical proved incremental properties of second-order HSIC to first-order HSIC in Proposition 4. 

\subsection{The tuning phase}

According to Proposition 4, the incremental information of second-order HSIC compared with first-order HSIC can identify the redundant edges based on a smaller second-order HSIC of variables shared child compared with first-order HSIC of variables with their child, the missing edges based on a larger second-order HSIC of variables shared child compared with first-order HSIC of variables with their child and the reverse edges based on a larger second-order HSIC between different variables and variables shared their child. In the tuning phase, we introduce the theoretical proved incremental principles to adjust the optimal structure obtained from the optimal phase by deleting the redundant edges, adding missing edges and constraining the output being a DAG. More concretely, we take three strategies by comparing  $\text{HSIC}({\rm X}_i,{\rm X}_j+{\rm X}_k)$ with $\text{HSIC}({\rm X}_i,{\rm X}_j)$ and $\text{HSIC}({\rm X}_i,{\rm X}_k)$ to learn a DAG more accurately. An outline of our tuning-phase algorithm is shown in Algorithm 1.

Before making any adjustments to tune the result given in the optimal phase, we need to adjust $\hat{\rm\bf W}$ obtained in the optimal phase to meet our specific requirement. Formally, we reshape $\hat{\rm\bf W}$ to a symmetric matrix with diagonal elements being zero such that $\hat{\rm\bf W} \in \left\{0,1\right\}^{d\times d}$ and $\hat{w}_{ii}=0$. The reason is that we could not decide dependent direction between variables only based on first-order HSIC. Therefore, we calculate second-order HSIC among variables and take three strategies to modify the graph following the principles provided by Proposition 4.

Deletion (Line 4 to 8 of Algorithm 1) : We compare each pair of variables with common child and delete the redundant edges between variables that the second-order of HSIC of variables shared child is smaller than their first-order HSIC with their child. Based on the principle proposed in Proposition 4(a), we can infer that $\text{ if } {\rm X}_j\in Pa({\rm X}_i) \text{ and }  {\rm X}_k \notin Pa({\rm X}_i)$:
\begin{equation}
\begin{aligned}
&{\rm HSIC}({\rm X}_j+{\rm X}_k,{\rm X}_i)\leq {\rm HSIC}({\rm X}_j,{\rm X}_i). \label{eq:13}
\end{aligned}
\end{equation}
Following Eq. (\ref{eq:13}), we recursively compare first-order HSIC to second-order HSIC for pair variables with common child and delete the edges by setting $\hat{w}_{ji}=-1,\hat{w}_{ki}=-1$ if ${\rm HSIC}({\rm X}_j+{\rm X}_k,{\rm X}_i)<\max \left\{{\rm HSIC}({\rm X}_j,{\rm X}_i),{\rm HSIC}({\rm X}_k,{\rm X}_i)\right\}$. The purpose of setting elements to -1 instead of 0 is that we could not figure which variable (${\rm X}_j,{\rm X}_k$) is not the parent of ${\rm X}_i$. So we need to delete both of edges temporarily and rejoin some of deleted edges in the addition step. 

Addition (Line 9 to 13 of Algorithm 1) : We compare each pair of variables with a common variable and add the missing edges between the variables that the second-order HSIC of variables shared child is larger than the largest of their first-order HSIC with their child. Based on the principle proposed in Proposition 4(b), we can infer that if ${\rm X}_j, {\rm X}_k \in Pa({\rm X}_i)$:
\begin{equation}
\begin{aligned}
&{\rm HSIC}({\rm X}_j+{\rm X}_s, {\rm X}_i) > \\
&\max \left\{ {\rm HSIC}({\rm X}_j,{\rm X}_i),{\rm HSIC}({\rm X}_s,{\rm X}_i)\right\}.\label{eq:14}
\end{aligned}
\end{equation}
Following Eq. (\ref{eq:14}), we iteratively compare first-order HSIC to second-order HSIC for pair of edges with common child which are not included in the learned graph and add both of edges by setting $\hat{w}_{ji}=1,\hat{w}_{ki}=1$ if ${\rm HSIC}({\rm X}_j+{\rm X}_s, {\rm X}_i) > \max \left\{ {\rm HSIC}({\rm X}_j,{\rm X}_i),{\rm HSIC}({\rm X}_s,{\rm X}_i)\right\}$. 

DAG-formalization (Line 14 to 17 of Algorithm 1) : We recursively find cycles and delete one of edges between the variables that the second-order HSIC of variables shared their child is the smallest. Based on the principle proposed in Proposition 4(c), we can infer that if ${\rm X}_j, {\rm X}_s \in Pa({\rm X}_i)$ and ${\rm X}_k \notin Pa({\rm X}_i)$:
\begin{equation}
{\rm HSIC}({\rm X}_j+{\rm X}_s,{\rm X}_i)>{\rm HSIC}({\rm X}_j+{\rm X}_k,{\rm X}_i). \label{eq:15}
\end{equation}
Following Eq. (\ref{eq:15}), we find loop and delete the edge with minimal second-order HSIC iteratively by setting $\hat{w}_{ki}=0$.
\begin{algorithm}[h]
\caption{Tuning-phase algorithm}
\begin{algorithmic}[1]
\STATE Input dataset $\rm\bf{X}$ and estimated adjacent matrix $\hat{\rm\bf{W}}$ of graph $\mathcal{G}$ in optimal phase

\STATE Calculate $\text{HSIC}({\rm X}_i,{\rm X}_j)$ for $i,j=1,...,d$ and $i\neq j$

\STATE Calculate $\text{HSIC}({\rm X}_i,{\rm X}_j+{\rm X}_k)$ for $i,j,k = 1,...,d$ and $i\neq j \neq k$

\FOR {each pair $\hat{w}_{ij} \text{ and } \hat{w}_{ik}$}
	\IF  {$\min(\text{HSIC}({\rm X}_i,{\rm X}_j), \text{HSIC}({\rm X}_i,{\rm X}_k)) \geq\text{HSIC}({\rm X}_i,{\rm X}_j+{\rm X}_k)$  for $\hat{w}_{ij}\neq 0 \text{ and } \hat{w}_{ik}\neq 0$}
		\STATE set $\hat{w}_{ij}=-1\text{ and }\hat{w}_{ik}=-1$
	\ENDIF
\ENDFOR
\FOR {each pair $\hat{w}_{ij} \text{ and } \hat{w}_{ik}$}
	\IF { $\max (\text{HSIC}({\rm X}_i,{\rm X}_j), \text{HSIC}({\rm X}_i,{\rm X}_k))<\text{HSIC}({\rm X}_i,{\rm X}_j+{\rm X}_k)$ for $\hat{w}_{ij}\neq 1 \text{ and } \hat{w}_{ik}\neq 1$}
		\STATE set $\hat{w}_{ij}=1\text{ and }\hat{w}_{ik}=1$
	\ENDIF
\ENDFOR
\WHILE {there exists loop in $\mathcal{G}$}
	\STATE find loop
	\STATE set $\hat{w}_{ij}=0$ with the minimal $\text{HSIC}({\rm X}_i,{\rm X}_j+{\rm X}_k)$ for node $i,j$ in the loop, where ${\rm X}_k \in Pa({\rm X}_i)$
\ENDWHILE
\STATE return $\hat{\rm\bf{W}}$
\end{algorithmic}
\end{algorithm}

\section{Numerical experiments}
In this section, we evaluate the performance of the algorithm compared with the state-of-the-art methods. For comparison, the following approaches are selected as baseline: GES \citep{Chickering2002}, NOTEARS (Linear) \citep{Zheng2018} for linear SEM, NOTEARS (nonparameter) \citep{Zheng2020} for nonlinear, NPVAR \citep{Gao2020} and CAM \citep{Buhlmann2014}.

\subsection{Synthetic data generation}
We construct the ground truth DAG on six generator models, comparing the performance in structure learning. In addition to those considered for the experimental analysis of the baseline algorithms \citep{Buhlmann2014,Zheng2020}, the more complex three generator models are considered, leveraging the nonlinearity of absolute value and mixing the different models. More concretely, the multilayer perceptron (MLP) model is used for the validation of the NOTEARS algorithm \citep{Zheng2020}, which is also applied to NOTEARS for approximating the generated functions between variables. The Tanh model is also used for the validation of the NOTEARS algorithm \citep{Zheng2020}, which is a single index model \citep{Sparse2013}. The Sigmoid Mix is used for the validation of the CAM algorithm, which adds a noise in the sigmoid function and differs from MLP. More three complex generator models are Abs using the sum of absolute value of variables's parents to generate the data, MLP-Tanh Mix mixing MLP and Tanh models and Abs-Tanh Mix mixing Abs and Tanh models to generate the data. We list the six generator models as follows:
\begin{enumerate}
\item MLP: ${\rm X}_i={\rm sigmoid}({\rm\bf PA(X_i)}{\rm\bf W1} ){\rm\bf W2}+e_i$ with ${\rm\bf W1}\in R^{a\times b}$ and ${\rm\bf W2}\in R^{b}$  where $a$ is the \\ number of  $X_i$'s parents and b is the number of layers, $w1_{ij},w2_{ij} \sim U([-2,-0.5]\cup [0.5,2])$;
\item Tanh: ${\rm X}_i=\text{tanh}(\sum_{{\rm X}_j\in{\rm\bf PA(X_i)}}\alpha{\rm X}_j)+e_i$ with $\alpha\sim U([-2,-0.5]\cup [0.5,2)$;
\item Sigmoid Mix: ${\rm X}_i={\rm sigmoid}(\sum_{{\rm X}_j\in{\rm\bf PA(X_i)}}\alpha{\rm X}_j+e_i)*\beta$, with $\alpha,\beta \sim U([-2,-0.5]\cup [0.5,2)$;
\item Abs: ${\rm X}_i=\sum_{{\rm X}_j\in{\rm\bf PA(X_i)}}\alpha\left|{\rm X}_j\right|+e_i$,with $\alpha \sim U([-2,-0.5]\cup [0.5,2)$;
\item MLP-Tanh Mix: ${\rm X}_i={\rm sigmoid}(\sum_{{\rm X}_j\in{\rm\bf PA(X_i)}}\alpha{\rm X}_j )*\beta+e_i$ with $\alpha,\beta \sim U([-2,-0.5]\cup [0.5,2)$ or ${\rm X}_i=\text{tanh}(\sum_{{\rm X}_j\in{\rm\bf PA(X_i)}}\alpha{\rm X}_j)+e_i$ with $\alpha \sim U([-2,-0.5]\cup [0.5,2)$;
\item Abs-Tanh Mix: ${\rm X}_i=\sum_{{\rm X}_j\in{\rm\bf PA(X_i)}}\alpha\left|{\rm X}_j\right|+e_i$,with $\alpha \sim U([-2,-0.5]\cup [0.5,2)$ or ${\rm X}_i=\text{tanh}(\sum_{{\rm X}_j\in{\rm\bf PA(X_i)}}\alpha{\rm X}_j)+e_i$ with $\alpha \sim U([-2,-0.5]\cup [0.5,2)$.
\end{enumerate}

MLP and Tanh models used for the validation of NOTEARS algorithm\citep{Zheng2020} can be found at https://github.com/xunzheng/notears, and Sigmoid Mix model used for the validation of the CAM algorithm can be found at https://github.com/bquast/ANM/tree/master/codeANM.

\subsection{Evaluation metrics}
We denote $T_p$ as the number of edges which are correctly recovered in the learned graph from the true graph, $F_n$ as the number of edges which are missing in the learned graph, and $F_p$ as the number of edges which are reversal and redundant in the real DAG. To evaluate performance among different approaches, we apply two different indicators for measuring the difference between the true graph and the learned graph.
\begin{enumerate}

\item  The structural intervention distance (SID) \citep{Peters}: SID not only counts the direct dependent relationship but also computes the path between all the pair of variables. Considering that missing edges $F_n$ is more harmful than incorrectly added edges $F_p$, SID gives the missing edges $F_n$ a larger weight, ranging in $[0, el+\hat{e}]$, where $e$ is the number of edges in the true graph, $\hat{e}$ is the number of edges in the learned graph, and $l$ is the number of edges in the longest path of all the pair of variables. The SID of the algorithm tends to 0 as the learned graph tends to the true graph.

 \item The area under the precision recall curve (AuPR): AuPR summaries the precision-recall curve and is the trade-off between precision $(T_p/(T_p+F_p))$ and recall $(T_p/(T_p+F_n))$, ranging in $[0,1]$. The AuPR of the algorithm tends to 1 as the learned graph tends to the true graph.

\end{enumerate}
SID and AuPR are calculated by python package cdt\footnote{https://fentechsolutions.github.io/CausalDiscoveryToolbox\\/html/causality.html}.

\subsection{Performance of the OT algorithm in the optimal and the tuning phases }
\begin{figure}[t]
	\centering
	\includegraphics[scale=0.25]{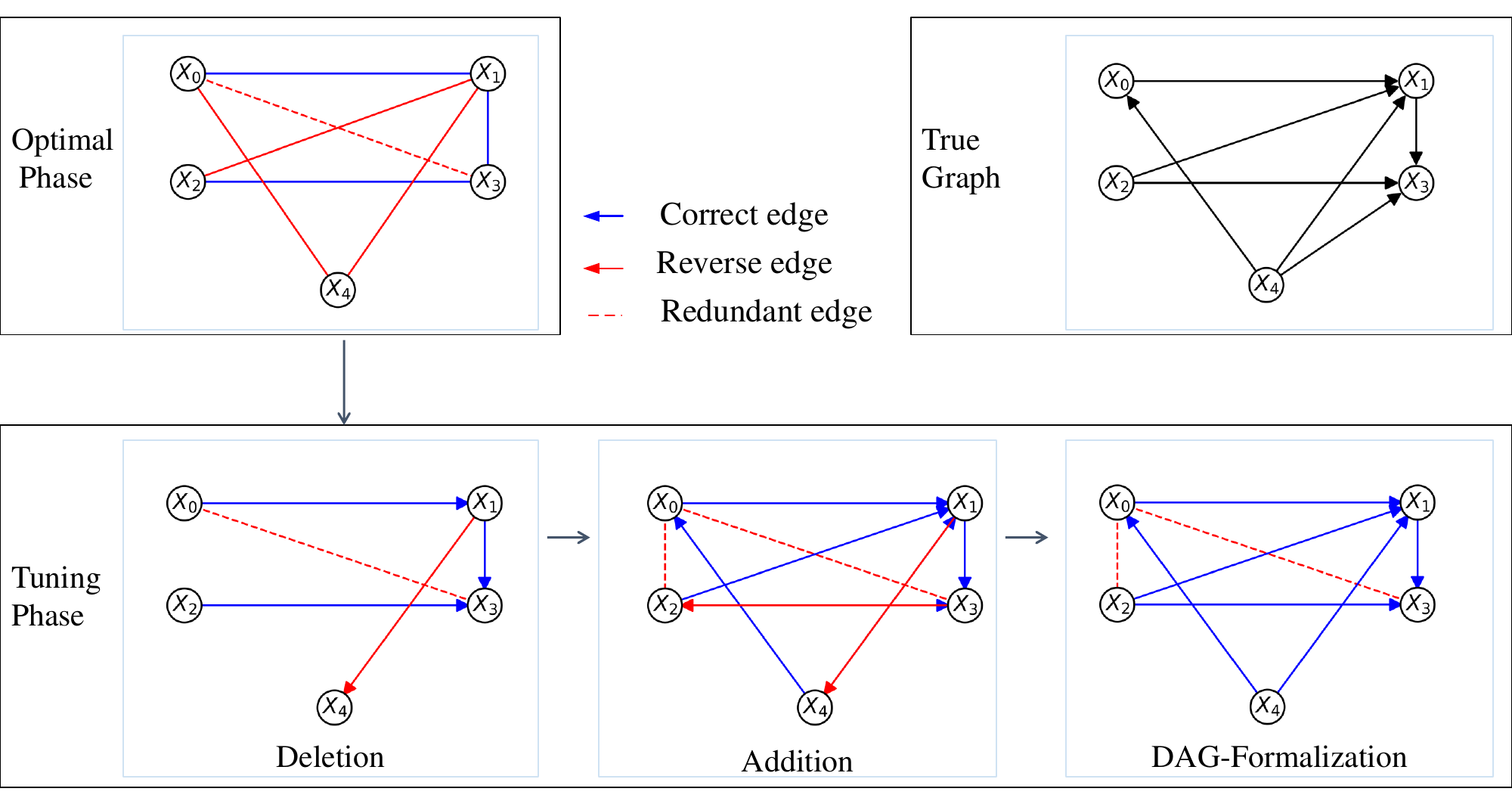}
	\caption{(Color online) The learned graphs of the OT algorithm in different phases. We give the true DAG in the upper right of figure for comparison. The blue arrow represents the correct edge, the red arrow represents the reverse edge, and the dashed red line represents the redundant edge. In optimal phase, the learned DAG contains five undirected edges. It can infer that the learned DAG in optimal phase is far from the true DAG, which only considers the first-order of HSIC based on Proposition 3. In the tuning phase, the algorithm locally amends the solution obtained from the optimal phase based on strategies of deletion, addition and DAG-formalization mentioned in Section IV. Three correct edges $\{{\rm X}_0\rightarrow {\rm X}_1, {\rm X}_1\rightarrow {\rm X}_3, {\rm X}_2\rightarrow {\rm X}_3\}$ are remained after deletion strategy. According to the learned DAG obtained in deletion strategy, three correct edges $\{{\rm X}_2\rightarrow {\rm X}_1,  {\rm X}_4\rightarrow {\rm X}_0, {\rm X}_4\rightarrow {\rm X}_1\}$ are added in addition strategy, while one reverse edge $\{{\rm X}_3\rightarrow {\rm X}_2\}$ and one redundant edge $\{{\rm X}_0 \rightarrow {\rm X}_2\}$ are added. Finally, two reverse edges $\{{\rm X}_1\rightarrow {\rm X}_4, {\rm X}_3\rightarrow {\rm X}_2\}$ are deleted based on DAG-formalization strategy and six correct edges $\{{\rm X}_0\rightarrow {\rm X}_1, {\rm X}_1\rightarrow {\rm X}_3, {\rm X}_2\rightarrow {\rm X}_1, {\rm X}_2\rightarrow {\rm X}_3, {\rm X}_4\rightarrow {\rm X}_0, {\rm X}_4\rightarrow {\rm X}_1\}$ are contained in the graph with two redundant edges $\{{\rm X}_0\rightarrow {\rm X}_2, {\rm X}_0\rightarrow {\rm X}_3\}$, one edge $\{{\rm X}_4\rightarrow {\rm X}_3\}$ is missed compared with true graph.}
	\label{fig:4}
\end{figure}

Here, we give an empirical analysis of the performance of  the OT algorithm in different phases. The true graph consists with 5 nodes and 7 edges, which is shown in Fig. \ref{fig:4}. We apply Abs-Tanh Mix model to the process of data generation. We show the learned graphs of the OT algorithm in different phases in Fig \ref{fig:4} and give the metrics of the OT algorithm in different phases in Table \ref{table:2}. 

As we can see in Fig. \ref{fig:4}, the learned graph in the optimal phase is far from the true graph, which contains one redundant edge and one missing edge. The main reason is that we only compare direct dependency between variables while some of dependency between variables are indirected. Then the algorithm locally amends the solution obtained from the optimal phase based on strategies of deletion, addition and DAG-formalization in the tuning phase. The final learned graph contains six correct edges $\{{\rm X}_0\rightarrow {\rm X}_1, {\rm X}_1\rightarrow {\rm X}_3, {\rm X}_2\rightarrow {\rm X}_1, {\rm X}_2\rightarrow {\rm X}_3, {\rm X}_4\rightarrow {\rm X}_0, {\rm X}_4\rightarrow {\rm X}_1\}$, two redundant edges $\{{\rm X}_0\rightarrow {\rm X}_2, {\rm X}_0\rightarrow {\rm X}_3\}$ and one edge $\{{\rm X}_4\rightarrow {\rm X}_3\}$ is missed compared with true graph. As is shown in Table \ref{table:2}. By deletion, the SID of the OT algorithm reduces to 11.0 and the AuPR of the OT algorithm increases to 0.6 compared with the one obtained in the optimal phase due to deleted edges are associated with redundant edges and reverse edges according to incremental property based on Eq. \ref{eq:13}. By addition, the Aupr of the OT algorithm increases to 0.72 and the SID of the OT algorithm increases to 15.0 due to added edges are reflected the dependence relationship between variables according to incremental property based on Eq. \ref{eq:14}, which are not constraint to be DAG. After DAG-formalization according to incremental property based on Eq. \ref{eq:15}, the learned graph is close to the true graph with the smallest SID and the largest AuPR.

\begin{table}
	\centering
	\caption{Metrics of learned graphs in different phases.}
	\begin{tabular}{l l l l}
		\hline
		\multicolumn{2}{c}{Phase} &SID&Aupr\\
		\hline
		Optimal phase&          &15.0&0.50\\
		\multirow{3}*{Tuning phase}& Deletion&11.0&0.60\\
							&Addition&15.0&0.72\\
							&DAG-formalization&$\rm\bf{1.0}$&$\rm\bf{0.82}$\\
		\hline
	\end{tabular}
	\label{table:2}
\end{table}

\subsection{Performance on synthetic data}

The five graphs used in this work are generated following the step as is mentioned in \cite{Kalisch2007a}, with the number of nodes $d \in \{10,20,25,30,40\}$,  the number of edges $s \in \{40,100,150,250,400\}$ and the size of data samples $n \in \{100,400,600,900,1600\}$. Note that we use six different generator models on these five graphs to distinguish the limitations of other methods and highlight the traits of our methods. In addition, the layers of MLP model during data generation is 100, denoting by 100-layers MLP model, which is consistent with the setting on NOTEARS.

\begin{figure*}[ht]\setcounter{subfigure}{0}
\centering
\subfloat
{\includegraphics[scale=0.11]{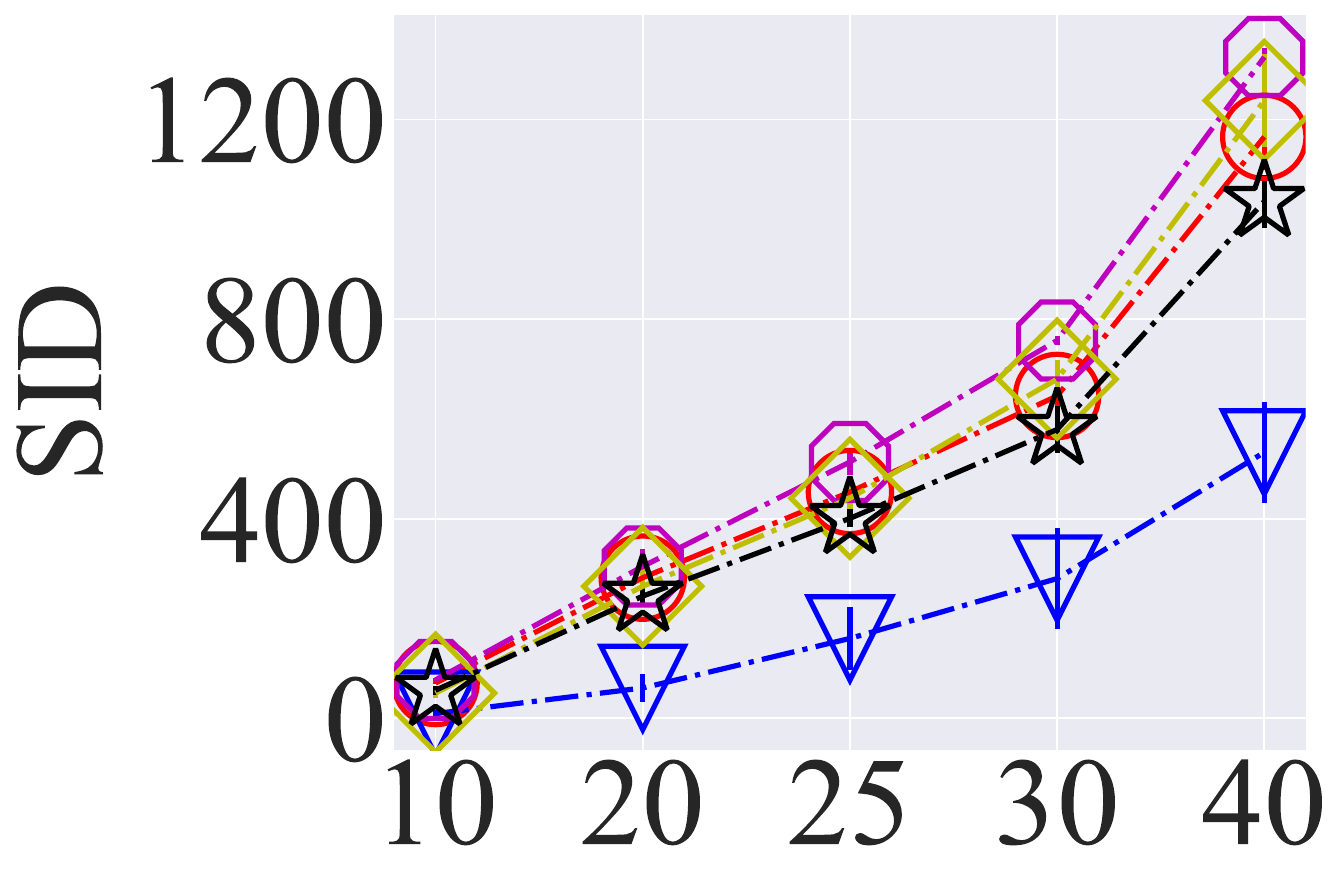}
}
\quad
\subfloat
{\includegraphics[scale=0.11]{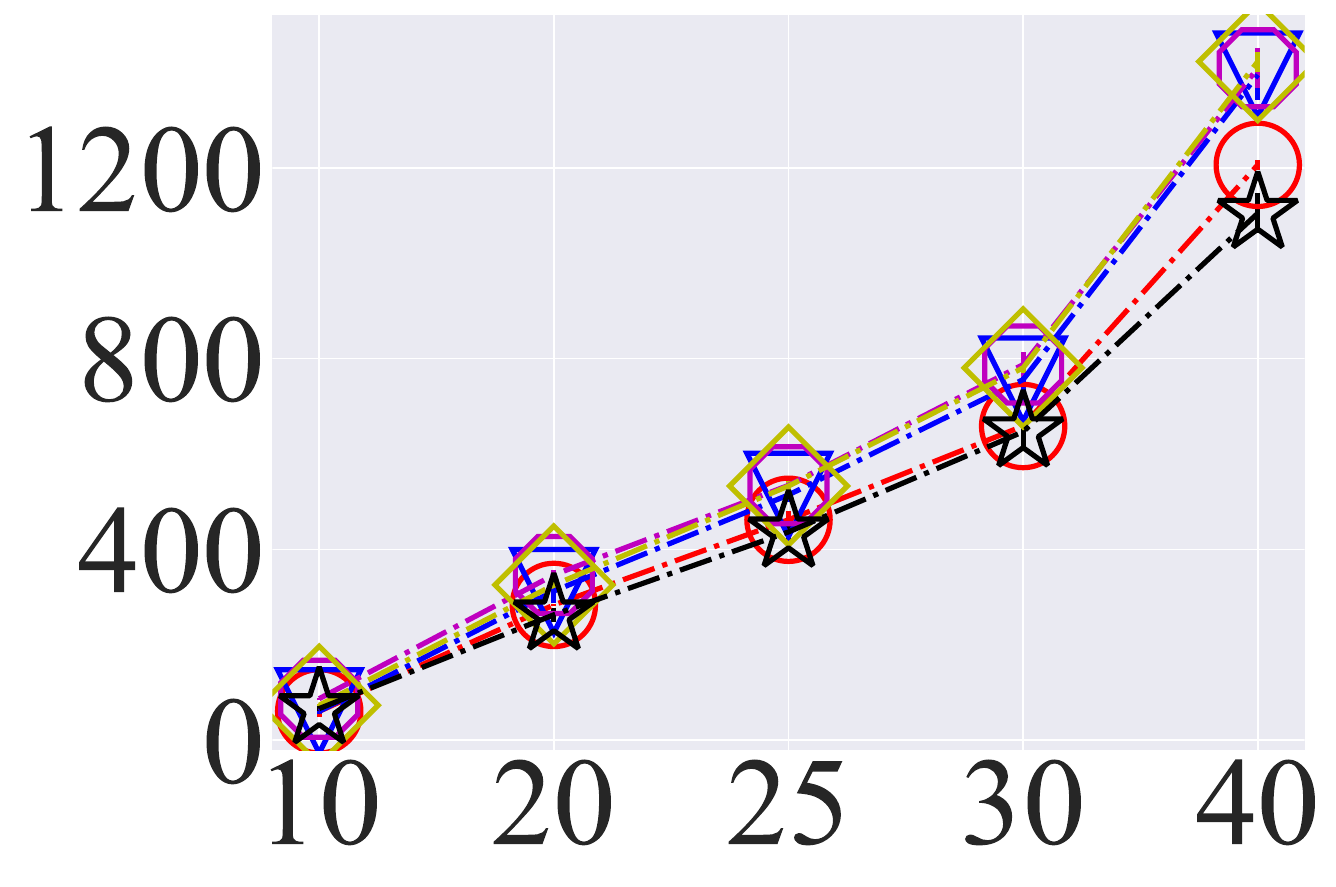}
}
\quad
\subfloat
{\includegraphics[scale=0.11]{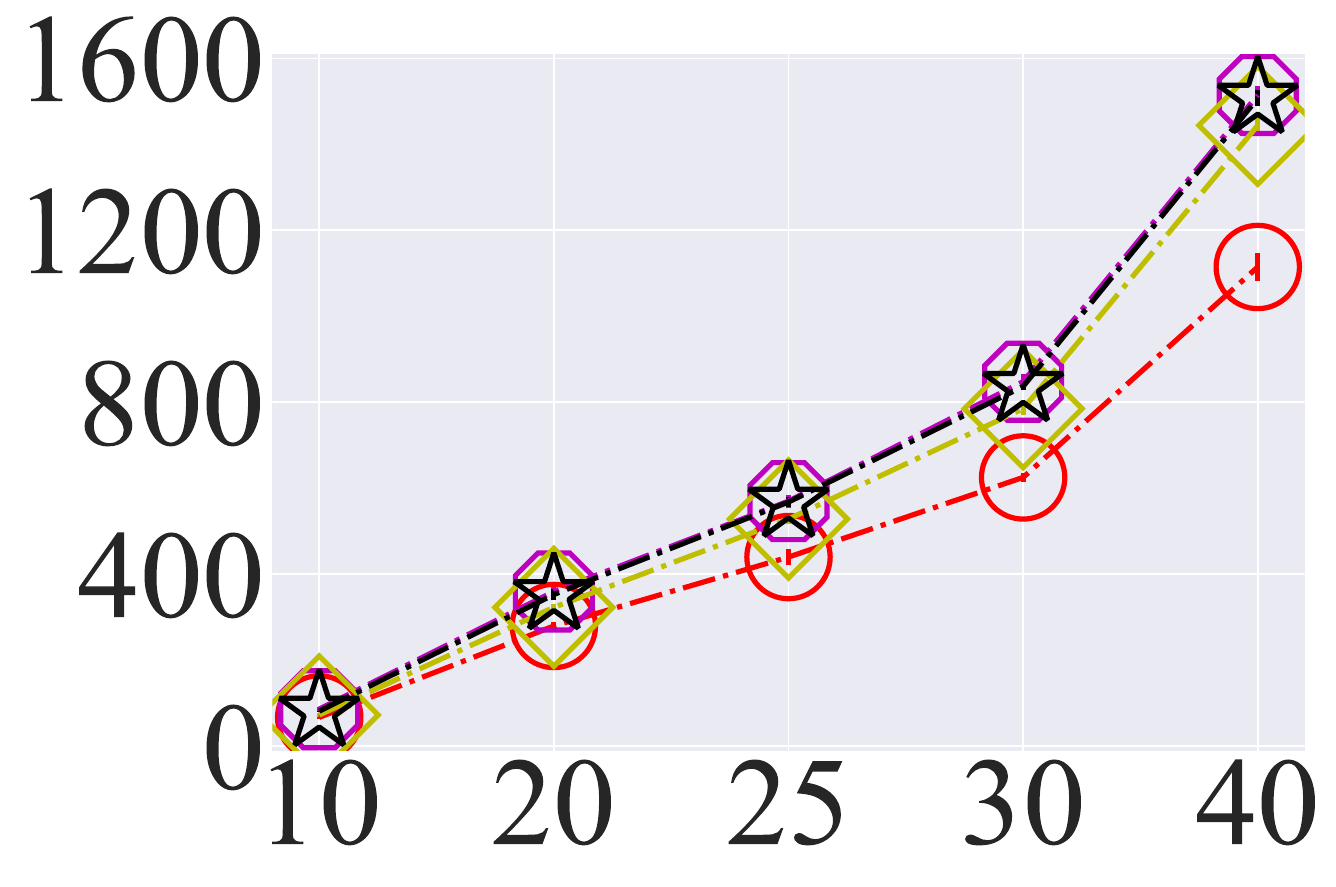}
}
\quad
\subfloat
{\includegraphics[scale=0.11]{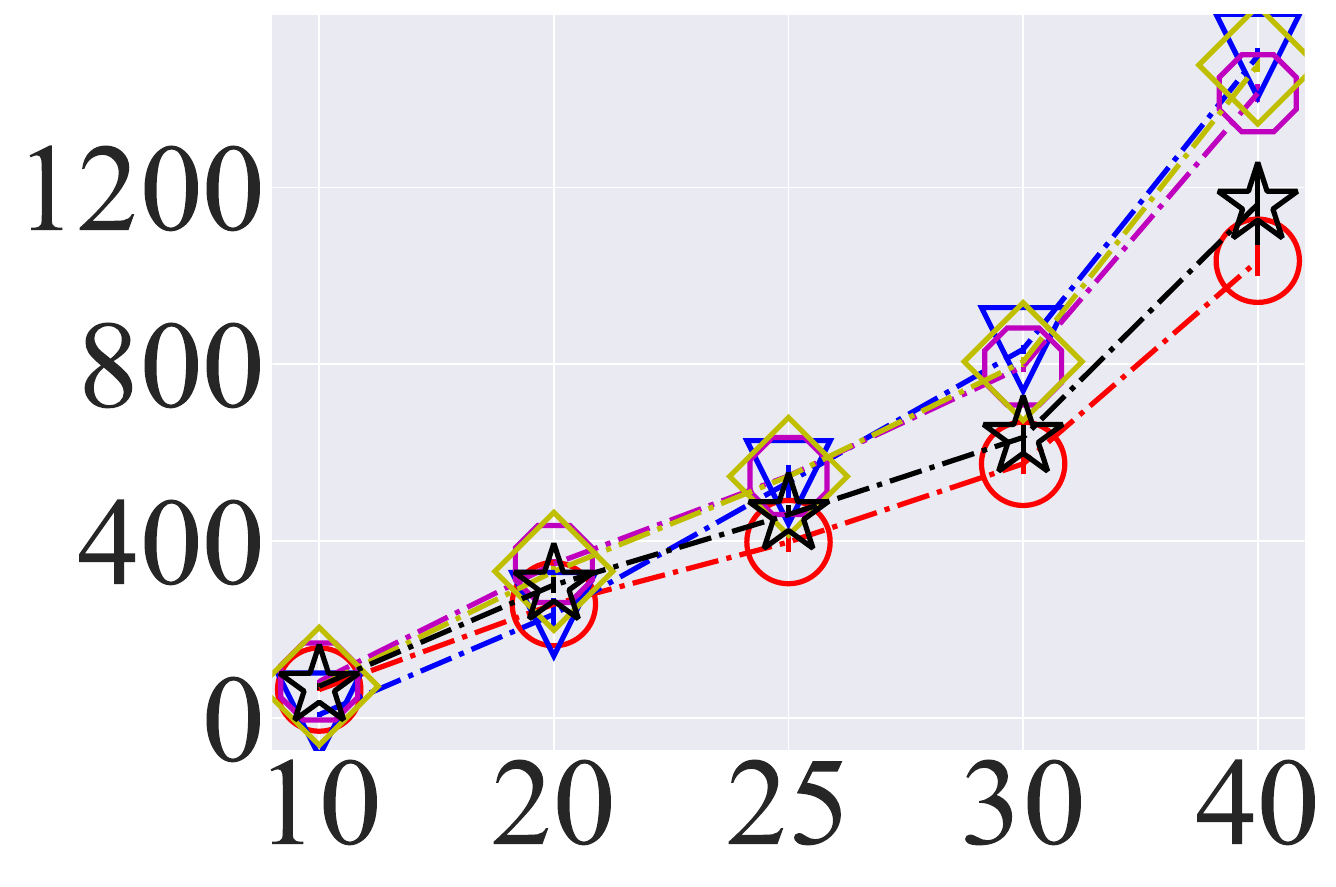}
}
\quad
\subfloat
{\includegraphics[scale=0.11]{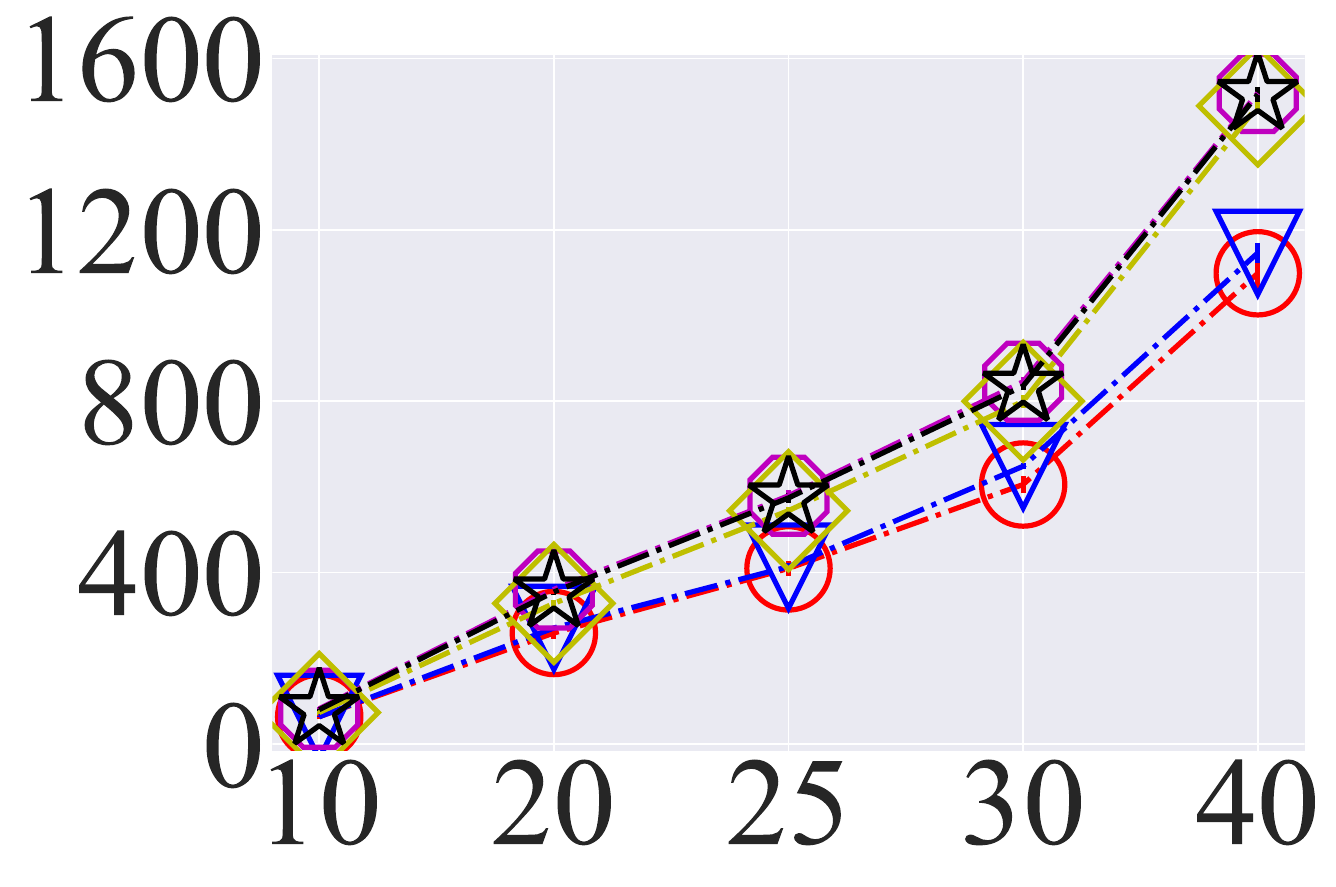}
}
\quad
\subfloat
{\includegraphics[scale=0.11]{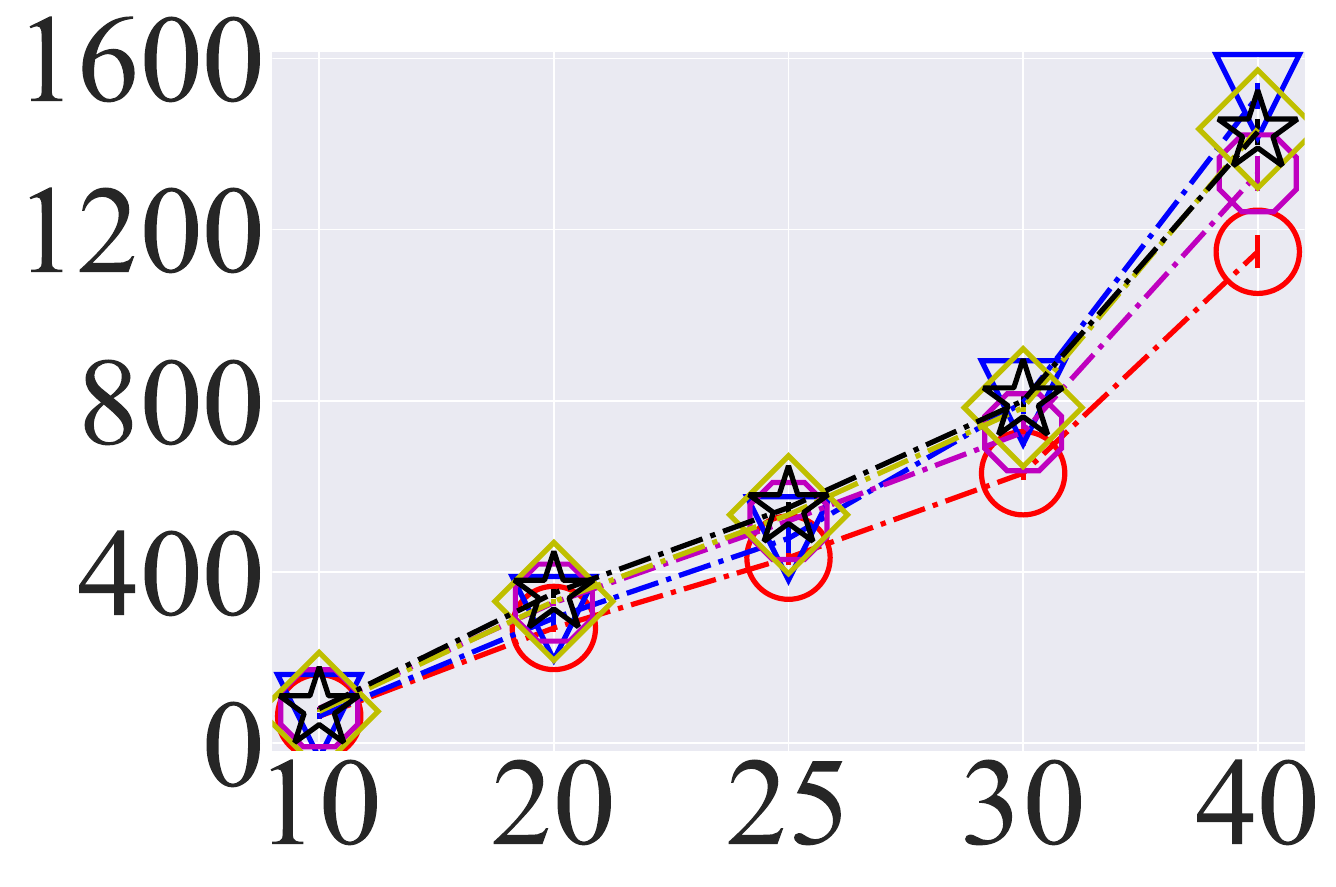}
}
\setcounter{subfigure}{0}
\subfloat[100-layers MLP]
{\includegraphics[scale=0.11]{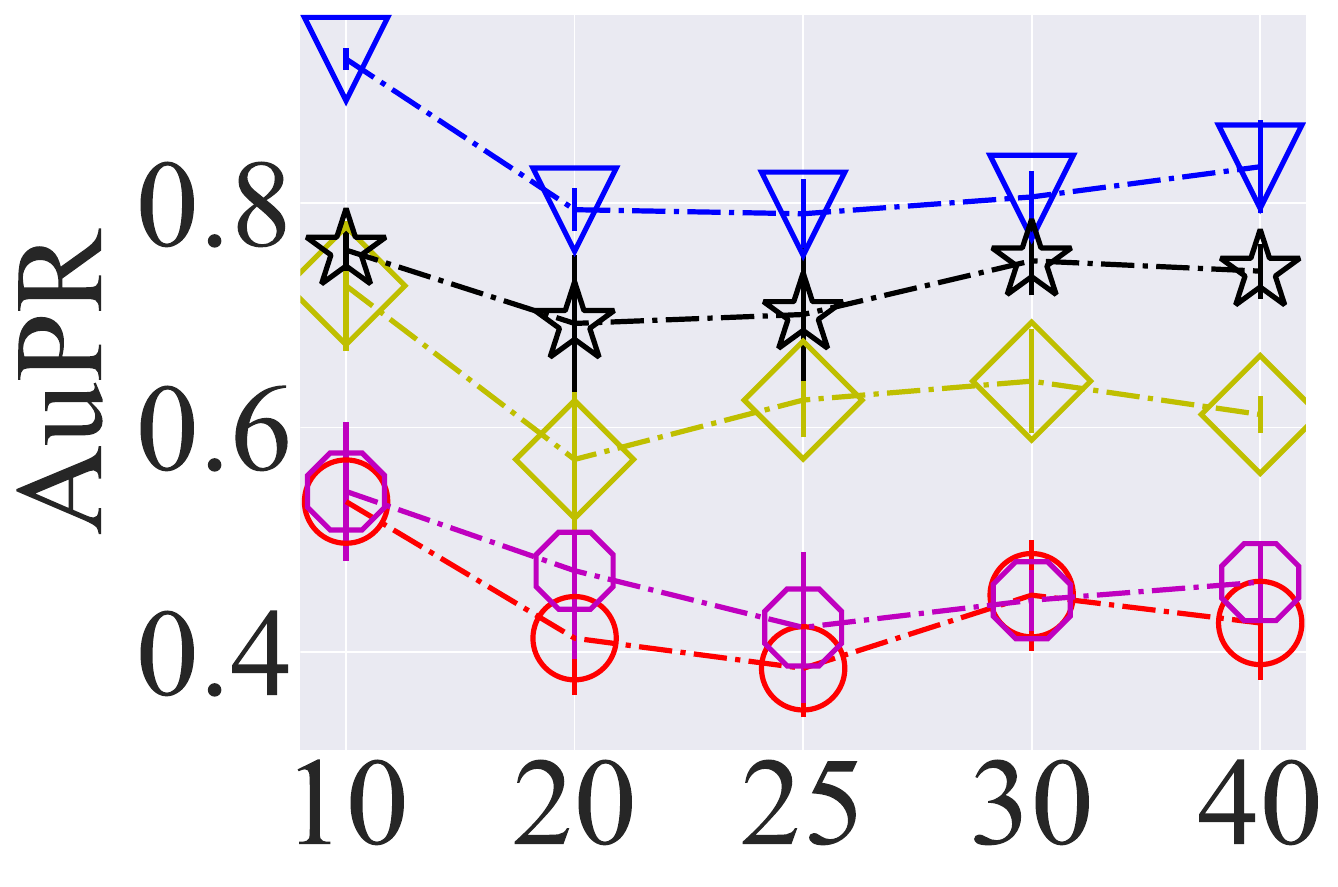}
}
\quad
\subfloat[Tanh]
{\includegraphics[scale=0.11]{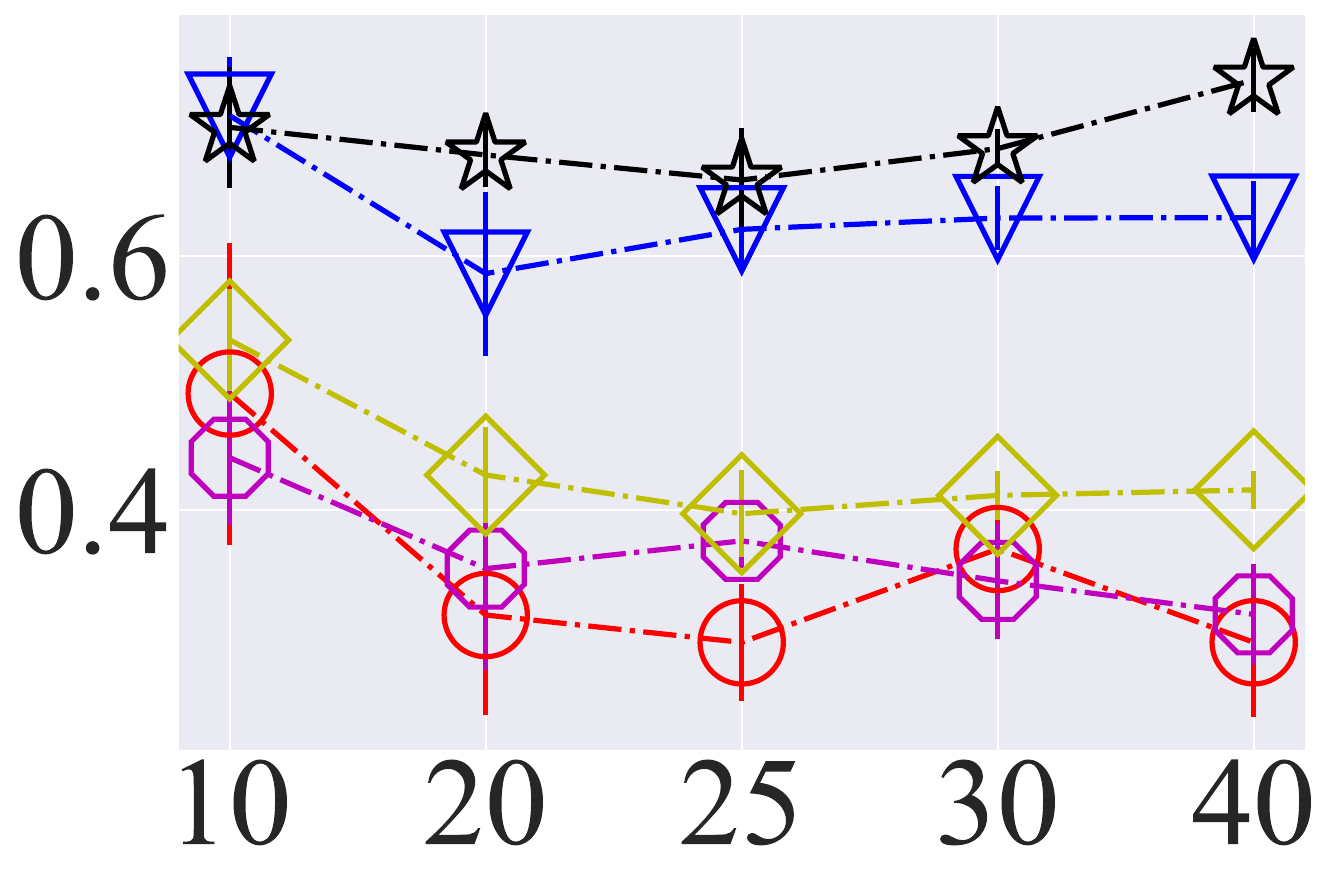}
}
\quad
\subfloat[Sigmoid Mix]
{\includegraphics[scale=0.11]{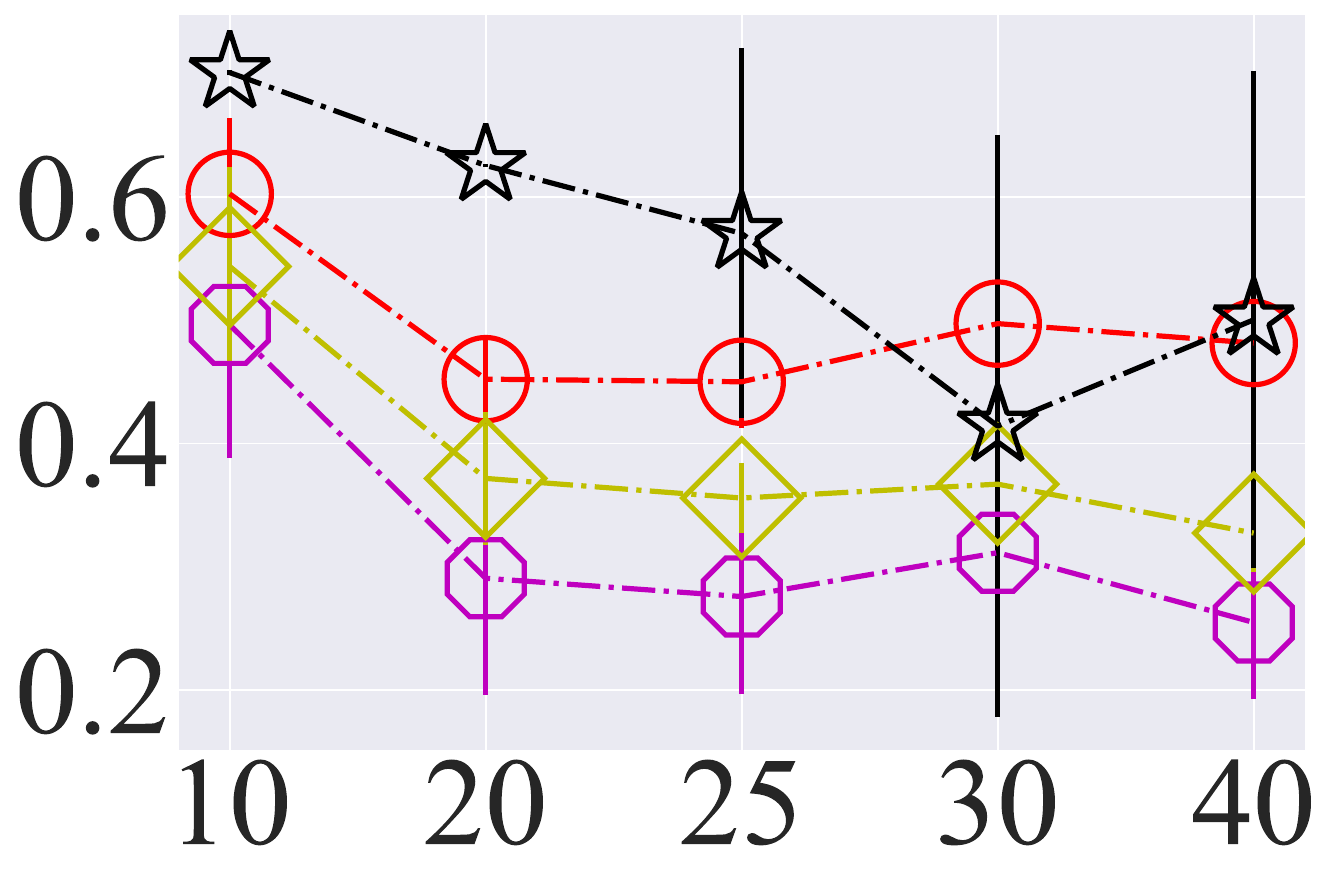}
}
\quad
\subfloat[Abs]
{\includegraphics[scale=0.11]{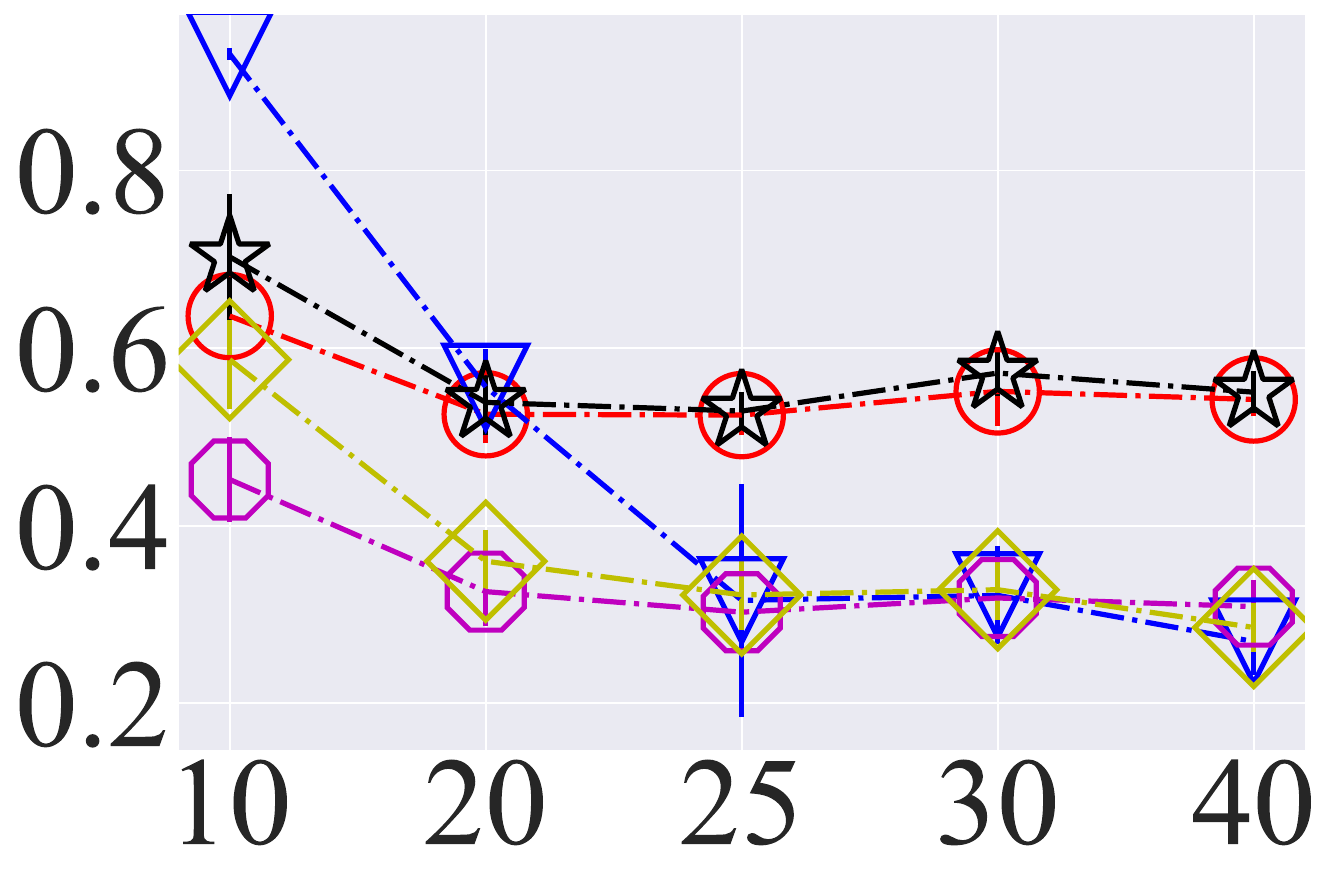}
}
\quad
\subfloat[MLP-Tanh Mix]
{\includegraphics[scale=0.11]{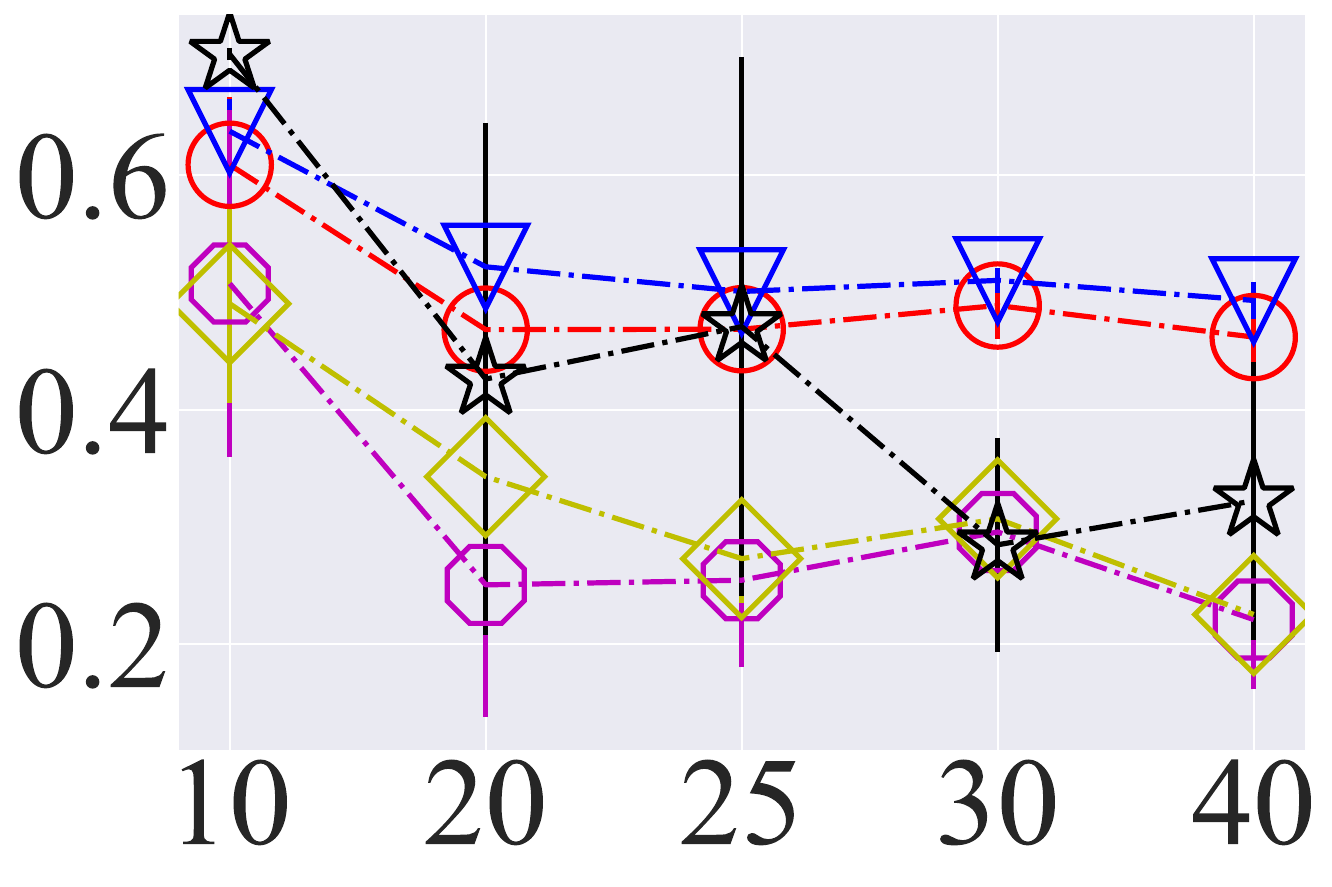}
}
\quad
\subfloat[Abs-Tanh Mix]
{\includegraphics[scale=0.11]{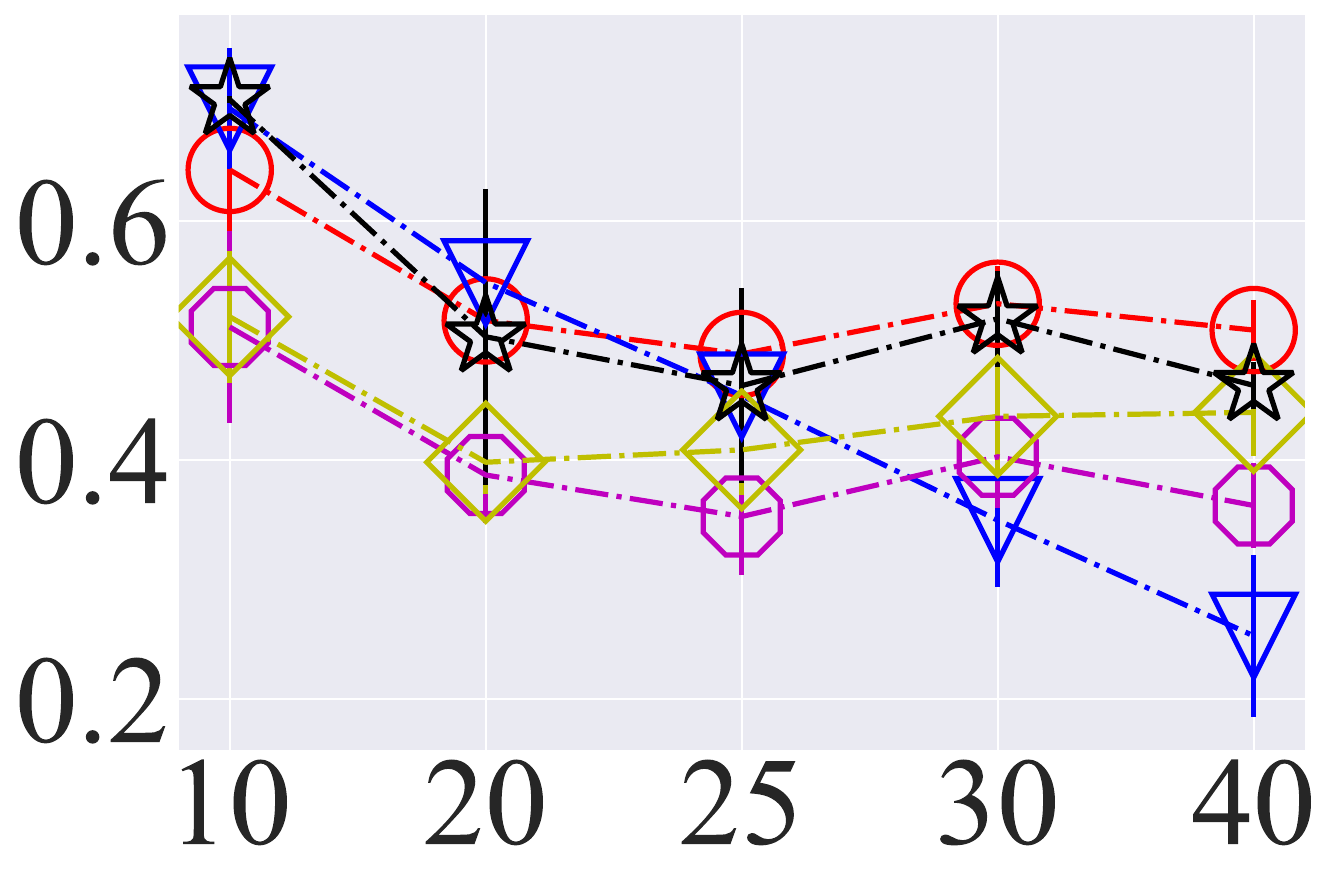}
}\quad
\subfloat
{\includegraphics[scale=0.1]{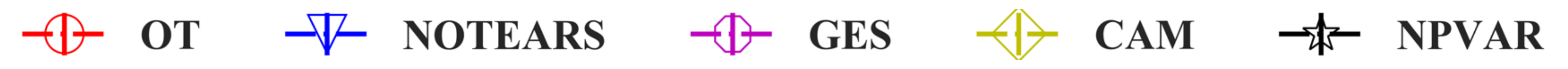}
}
\caption{(Color online) Performance of structure learning methods measured by SID (the value tends to 0 as the learned graph tends to the true graph) and AuPR (the value tends to 1 as the learned graph tends to the true graph) to ground truth on graphs with different generator mechanisms. When the size of graph $d=40$ in Sigmoid Mix model, the SID of the OT algorithm is 1114.5, while the SID of NPVAR is 1507.4, CAM is 1444.2 and GES is 1514.8. When the size of the graph $d=40$ in Abs model, the SID of the OT algorithm is 1034, while the SID of NPVAR is 1162, GES is 1412.6, CAM is 1476.5 and NOTEARS is 1497.1. When the size of the graph $d=40$ in MLP-Tanh Mix model, the SID of the OT algorithm is 1098.4, while the SID of NOTEARS is 1146, CAM is 1489, NPVAR is 1515.8 and GES is 1519. When the size of the graph $d=40$ in Abs-Tanh Mix model, the SID of the OT algorithm is 1148.2, while the SID of GES is 1331.4, NPVAR is 1428, CAM is 1435 and NOTEARS is 1511.8. Overall, the proposed OT algorithm generates the smallest SID, especially when the data generator models are Sigmoid Mix, Abs, MLP-Tanh Mix or Abs-Tanh.}
\label{fig:5}
\end{figure*}

We evaluate the performance of the proposed algorithm and state-of-art approaches GES, CAM, NOTEARS and NPVAR on synthetic data. Figure \ref{fig:5} shows SID (the value tends to 0 as the learned graph tends to the true graph) and AuPR (the value tends to 1 as the learned graph tends to the true graph) of the OT algorithm compared with different algorithms in various settings.

Overall, the proposed OT algorithm attains the smallest SID across a wide range of settings, especially when the data generator models are Sigmoid Mix, Abs, MLP-Tanh Mix or Abs-Tanh. It also can be observed that the SID and AuPR of the OT algorithm stay stable for different data generator models, as it does not make assumptions on the dependent relationship between variables but uses a sumable formation to approximate the dependent relationship between variables based on Proposition 2. 

When the size of graph $d=40$ in Sigmoid Mix model, the SID of the OT algorithm is 1114.5 and is the smallest compared with other baseline algorithms,  while the SID of CAM is 1444.2, NPVAR is 1507.4 and GES is 1514.8. On the other hand, the AuPR of the OT algorithm is 0.5 and is equal to the AuPR of NPVAR, which are the largest compared with other algorithms, while the AuPR of CAM is 0.3 and GES is 0.25. When the size of the graph $d=40$ in Abs model, the SID of the OT algorithm is 1034 and is the smallest compared with other baseline algorithms, while the SID of NPVAR is 1162, GES is 1412.6, CAM is 1476.5 and NOTEARS is 1497.1. On the other hand, the AuPR of the OT algorithm is 0.5 and is equal to NPVAR, which are the largest compared with other algorithms, while the AuPR of GES is 0.3, same as CAM and NOTEARS. When the size of the graph $d=40$ in MLP-Tanh Mix model, the SID of the OT algorithm is 1098.4 and is the smallest compared with other baseline algorithms, while the SID of NOTEARS is 1146, CAM is 1489, NPVAR is 1515.8 and GES is 1519. On the other hand, the AuPR of the OT algorithm is 0.5 and is equal to NOTEARS, which are the largest compared with other algorithms, while the AuPR of NPVAR is 0.3, CAM and GES are 0.2. When the size of the graph $d=40$ in Abs-Tanh Mix model, the SID of the OT algorithm is 1148.2 and is the smallest compared with other baseline algorithms, while the SID of GES is 1331.4, NPVAR is 1428, CAM is 1435 and NOTEARS is 1511.8. On the other hand, the AuPR of the OT algorithm is 0.5 and is equal to NPVAR, which are the largest compared with other algorithms, while the AuPR of CAM and GES are 0.4 and NOTEARS is 0.3. Note that the performance of the OT algorithm on the SID indicator is better than its performance on the AuPR indicator. As we mentioned in Subsection V B, SID gives the missing edges $F_n$ a larger weight compared with AuPR as missing edges are associated with the true graph. The main reason for the outperforming of the OT algorithm on the SID indicator is that compared with the AuPR indicator, the SID indicator gives missing edges a larger weight, validating that the OT algorithm has the ability to reccover more edges that the baseline algorithms miss. Moreover, the experiments show that NOTEARS does not work for Sigmoid Mix model due to numerical issues. Compared to the overall performance, the OT algorithm obtains an advantage in Sigmoid Mix, Abs, MLP-Tanh Mix or Abs-Tanh models.

From Fig. \ref{fig:5}(a), one can find that the performance obtained by the NOTEARS algorithm with 100-layer MLP is better than the one given by the OT algorithm. The reason lies in the fact that NOTEARS specifically employ MLP to approximate the dependent relationship between variables, which has an advance in the generator model using 100-layers MLP. On the other hand, the performance of NOTEARS is significantly outperforms than other algorithms compared its performance on other models. When the size of the graph $d=40$ in the 100-layers MLP model, the SID of NOTEARS is 533, while the SID of NPVAR is 1036, the OT algorithm is 1165, CAM is 1238.2 and GES is 1325.2. As is mentioned in the theoretical analysis in Section III, the main idea behind the OT algorithm is to identify the underlying graph by comparing different order of dependence level among variables given the determined parents set, which does not rely on the form of generator models, while 100 layers transformation on variables would give more extra dependent information to NOTEARS as NOTEARS uses the MLP to approximate the dependent relationship between variables. Thus, we set the layers of MLP model to 2 layers to verify whether the 100 layers transformation on variables would give more extra dependent information to NOTEARS. Table \ref{table:6} shows the performance of the OT algorithm on the 2-layers MLP model with different algorithms in various settings. When the size of the graph $d=40$ in the 2-layer MLP model, the SID of the OT algorithm is 1241.1 and is the samllest compared with other baseline algorithms, while the SID of NPVAR is 1253.8, NOTEARS is 1355.8, GES is 1394.1 and CAM is 1413.2. It can be found that the performance of NOTEARS on the 2-layers MLP is dominated by the OT algorithm, which indicates that the performance of NOTEARS depends on the dependent relationship between variables.

\begin{table}
	\centering
	\caption{ The performance of the OT algorithm with the 2-layers MPL model compared with state-of-art approaches GES, CAM, NOTEARS and NPVAR measured by SID.}
	\begin{tabular}{l l l l l l }
		\hline
		{} &OT&NOTEARS&NPVAR&CAM&GES\\
		\hline
		$d=10$ &$\rm\bf{68.1}$&62.7&71.8&72.1 &83.5\\
		$d=20$&$\rm\bf{284.8}$&294.5&310.3&316.1&333.9\\
		$d=25$& $\rm\bf{460.4}$&\rm\bf{443.6}&469.8&522.1&533.2\\
		$d=30$& $\rm\bf{676.3}$&742.7&726.8&765.1&799.8\\
		$d=40$& $\rm\bf{1241.1}$&1355.8&1253.3&1413.2&1394.1\\
		\hline
	\end{tabular}
	
	\label{table:6}
\end{table}

We note that the OT algorithm is dominated by NPVAR in Tanh model in Fig. \ref{fig:5}(b). Specially, the performance of the OT algorithm declines in Tanh model compared its performance on the other models. The SID of NOTEARS is 1104.5, while the SID of the OT algorithm is 1206, NOIEARS is 1394.9, GES is 1408.7 and CAM is 1422.5 when the size of graph $d = 40$. On the other hand, the AuPR of NPVAR is 0.7, while the AuPR of NOTEARS is 0.6, CAM is 0.4, GES is 0.3 and the OT algorithm is 0.3. The main reason is that the Gaussian kernel in HSIC, which is used in the OT algorithm to measure the dependence level between variables, can not capture the deviation generated by sigmoid function. Thus, we use the sigmoid kernel $k({\rm X}_i,{\rm X}_i')={\rm tanh}(\sigma {\rm X}_i^T{\rm X}_i'+\gamma)$ to replace the Gaussian kernel in HSIC to examine the performance of the OT algorithm in nonlinear model, where $\sigma=1$ and $\gamma=0$ according to \cite{Lin2003}. Table \ref{table:5} shows the performance of the OT algorithm with sigmoid kernel in SID and AuPR compared with different algorithms in various settings in Tanh model. When the size of graph $d=40$, the OT algorithm obtains the smallest SID as 740 compared with state-of-art approaches GES, CAM, NOTEARS and NPVAR. The experiments show that when replacing the Gaussian kernel with a sigmoid kernel, the OT algorithm outperforms in Tanh model, which suggestss that a sigmoid kernel in HSIC does capture the deviation in Tanh model.

\begin{table}
	\centering
	\caption{ The performance of the OT algorithm with sigmoid kernel in Tanh model compared with state-of-art approaches GES, CAM, NOTEARS and NPVAR measured by SID and AuPR. }
	\begin{tabular}{l l l l l l l}
		\hline
		\multicolumn{2}{c}{} &OT&NOTEARS&NPVAR&CAM&GES\\
		\hline
		\multirow{2}*{$d=10$}&SID &$\rm\bf{37}$&60.4&66.3 &73.3 &87\\
						 &AuPR&{\rm\bf 0.8}&0.7&0.7&0.5&0.4\\		
		\multirow{2}*{$d=20$}& SID &$\rm\bf{170.2}$&312.5&262.8&325.3&346.3\\
						&AuPR&{\rm\bf 0.7}&0.6&{\rm\bf 0.7}&0.4&0.4\\
		\multirow{2}*{$d=25$}& SID& $\rm\bf{279}$&514.1&436.5&532.8&534.2\\
						&AuPR&{\rm\bf0.7}&0.6&0.7&0.4&0.4\\
		\multirow{2}*{$d=30$}& SID& $\rm\bf{422.2}$&755.9&646.5&780.5&787.3\\
						&AuPR&{\rm\bf0.7}&0.6&{\rm\bf0.7}&0.4&0.3\\
		\multirow{2}*{$d=40$}& SID& $\rm\bf{740}$&1394.9&1104.5&1422.5&1408.7\\
						&AuPR&{\rm\bf0.7}&0.6&{\rm\bf0.7}&0.4&0.3\\
		\hline
	\end{tabular}
	
	\label{table:5}
\end{table}

The performance of the OT algorithm stays stable for each same graph with different dependent mechanisms, as it does not need to learn DAGs through estimating dependent function among variables and adjust the optimal solution regarding the difference between direct dependence and indirect dependence in the tuning phase, as is shown in Fig. \ref{fig:5}. On the other hand, when the ground truth is Tanh model, the performance of the OT algorithm obtains an advantage with sigmoid kernel function used in HSIC, as is shown in Table \ref{table:5}. In addition, the experiments show that NOTEARS can not learn the DAGs in Sigmoid Mix model, while the OT algorithm obtians a stable advantage in six generator models.
\subsection{Performace on real-world data}

Finally, we evaluate the performance of the OT algorithm compared with other baseline algorithms for two real datasets.The first real dataset is from \cite{Sachs2005} that is a common benchmark in Bayesian network models as it comes with a well-known consensus graph. This dataset consists of human immune system molecules ($n=7466$, $d=11$ with 20 edges) with continuous measurement on signaling of multiple phosphorylated protein and phospholipid, which is widely adopted by the biological community.

Table \ref{table:3} shows the performance of structure learning methods measured by SID (the value tends to 0 as the learned graph tends to the true graph) and AuPR (the value tends to 1 as the learned graph tends to the true graph), from which one can find that the largest AuPR is obtained by the OT algorithm. The SID of NPVAR is the smaller than the OT algorithm but the number of edges learned from NPVAR is 43 and larger than the one obtained by the OT algorithm, which suggests that the learned graph obtained by NPVAR contains more redundant edges compared with the OT algorithm with the largest AuPR.

\begin{table}
	\centering
	\caption{Metrics of estimated graph for two real datasets. The largest AuPR is obtained by the OT algorithm both in Protein graph and House price graph. For Protein graph, the SID of NPVAR is the smaller than the one obtained by the OT algorithm but the number of edges learned from NPVAR is 43 and larger than the one obtained by the OT algorithm, which suggests that the learned graph obtained by NPVAR contains more redundant edges compared with the OT algorithm with the largest AuPR. For House price graph, the AuPR of GES is equal to the OT algorithm but the SID of GES is 51 and larger than the OT algorithm, which indicates that the estimated graph learned by GES are missing some edges while the learned graph obtained from the OT algorithm has less missing edges compared with the true graph. In addition, the CAM algorithm does not work for House price graph due to the regressor of the model that does not converge on dicrete variables.}

\begin{tabular}{ l l l l l l l}
		\hline
		\multirow{3}*{Methods} &\multicolumn{3}{c}{Protein graph}&\multicolumn{3}{c}{Hous price graph}\\
		\\
		
						  &SID&AuPR&Edges&SID&AuPR&Edges\\
		\hline
		OT                     &${\rm\bf 86.0}$&${\rm\bf 0.4}$&${\rm\bf 10.0}$&{\rm\bf 28.0}&{\rm\bf 0.4 }&{\rm\bf 41}\\
		NOTEARS 			 &88.0&0.2&19.0&69.0&0.2&45\\
		NPVAR      		 &$\rm\bf{82.0}$&0.3&43.0&66.0&0.1&43\\
		CAM        		 &87.0&0.3&27.0&-&-&-\\
		GES          		 &95.0&0.3&36.0&51&{\rm \bf 0.4}&32\\
		\hline
	\end{tabular}
	\label{table:3}
\end{table}
The second well-studied house price graph is associated with modeling sale prices as a function of the property attributes, which is first used by De Cock\citep{DeCock2011} as a tool for statistics education. According to Conrady and Jouffe \citep{Conrady2015}, the dataset includes 11 variables with 12 edges, that is, $n=2930$, $d=11$ with $E=12$ edges for dateset.

Table \ref{table:3} shows the performance of structure learning methods measured by SID (the value tends to 0 as the learned graph tends to the true graph) and AuPR (the value tends to 1 as the learned graph tends to the true graph), from which one can find that the largest AuPR and the smallest SID are obtained by the OT algorithm. One could also notice that the AuPR of the GES equals to the one obtained by the OT algorithm while the SID of GES is 51, which is larger than the one obtained by the OT algorithm, indicating that the estimated graph learned by the GES misses some edges while the learned graph obtained by the OT algorithm has less missing edges compared with the empirical graph. In addition, the CAM algorithm does not works for House price graph due to the regressor of the model that does not converge on dicrete variables.

\section{Conclusion}
Learning DAG from data is an NP-hard problem. The DAG discovery methods can be sub-categorized to combinatoric graph search and continuous optimization by finding the most dependent component of variables. Although the trade-off between space search and validity improve the efficiency for learning DAG, the score-based metheds focus on the global optimization regardless of indirect dependence where the local variables may have direct and indirect dependence simultaneously, leading to missing edges. In this paper, we present an identifiability condition based on a determined subset of parents for ANMs to recover the underlying DAG by comparing different order of dependent level between variables. Then we theoretically prove three properties of high-order HSIC. The first property is to to identify the redundant edges based on a smaller second-order HSIC of variables shared child compared with first-order HSIC of variables with their child. The second property is to identify the missing edges based on a larger second-order HSIC of variables shared child compared with first-order HSIC of variables with their child. The third property is to identify the reverse edges based on a larger second-order HSIC between different variables and variables shared their child. Based on the theoretically proved properties, we develop a two-phase algorithm, namely OT algorithm, to identify the direct dependent from indirect dependent. In the optimal phase, the OT algorithm learns an optimal structure based on the first-order HSIC (Proposition 3), which uses Gumbel-softmax trick to transform discrete learning of first-order HSIC to a continuous optimization. In the tuning phase, the OT algorithm introduce the theoretically proved properties of high-order HSIC to adjusts the optimal structure using deletion, addition and DAG-formalization strategies (Proposition 4). 

We measure the performance of the proposed algorithm OT comparing with state-of-art approaches GES, CAM, NOTEARS and NPVAR on synthetic data based on MLP, Tanh, Sigmoid Mix, Abs, MLP-Tanh Mix and Abs-Tanh Mix models. Overall, the proposed OT algorithm generates the smallest SID across a wide range of settings, especially when the data generator models are Sigmoid Mix, Abs, MLP-Tanh Mix or Abs-Tanh. The performance of the OT algorithm on the SID measurement is better than the one generated by the AuPR measurement. As we mentioned in Subsection V B, SID gives the missing edges $F_n$ a larger weight compared with AuPR as missing edges are associated with the true graph. The main reason for the outperformance of the OT algorithm on the SID measurement lies in the fact that compared with the AuPR measurement, the SID measurement gives missing edges a larger weight, validating that the OT algorithm has the ability on recovering more edges that the baselined algorithms miss. In addition, the performance of NOTEARS on the 2-layers MLP is dominated by the OT algorithm, which suggests that the performance of NOTEARS depends on the dependent relationship between variables. The experiments show that when replacing the Gaussian kernel with a sigmoid kernel, the OT algorithm obtains an advantage in Tanh model, especially being clearer as the size of graph increases, which indicates that a sigmoid kernel in HSIC does capture the deviation in Tanh model. Our experiments for two real datasets show that the OT algorithm is effective and accurate for discovering DAGs and outperforms four classical algorithms. Especially in House price graph, the SID of the OT algorithm is 41 smaller than the one obtained by NOTEARS.

To tune the result given in the optimal phase, we delete the redundant edges, add missing edges and constraint the output being a DAG according to the principle provided by Proposition 3 and take three steps by comparing  $\text{HSIC}({\rm X}_i,{\rm X}_j+{\rm X}_k)$ with $\text{HSIC}({\rm X}_i,{\rm X}_j)$ and $\text{HSIC}({\rm X}_i,{\rm X}_k)$ to determine the direction of the edges and correct the misestimated edges. Although we do not rely on model assumptions to tune the optimal result, the comparison on independence between high-order HSIC is approximated by additive functions, which lowers the precision of the algorithm. Hence, it is an important problem to find a general function to characterize the high-order HSIC and should be investigated in the future. Another extension is to give the theoretical analysis of how kernel function used in HSIC works on data with different casual functions, as our empirical experiments show that the OT algorithm only obtains an advantage in Tanh model when using a sigmoid kernel function. Furthermore, the tuning phase of the algorithm can be fruitfully applied to study different structures that go beyond three variables, for which efficiency and accuracy would be expected to promote.
\section*{Acknowledgments}
We would like to acknowledge support for this project
from the National Natural Science Foundation of China (Grant Nos. 71771152, 71632006, 72032003, 82041020).


\section*{Appendix A}
{
\setcounter{equation}{0}
\renewcommand\theequation{A.\arabic{equation}} 
\subsection*{A.1 Proof of Theorem 1}
\noindent
{\bf Proof}. \sloppy Without loss of generality, let $PA_d({\rm X}_i)$ be the determined parents and $PA({\rm X}_i)$ be all parents for each variable $X_i$ for $i=1,...,d$. Now, we prove the identifiability of ANMs using mathematical induction, as follows:

Step (1): Suppose there are at least one determined parent for variable ${\rm X}_i$, we have :
\begin{equation}
\begin{aligned}
&dep({\rm X_i},PA_d({\rm X}_i)))\\
&=\sum_{h,h'}|{\rm E}(h(PA({\rm X}_i))+e_{i})h'(PA_d({\rm X}_i))\\
&-{\rm E}(h(PA({\rm X}_i))+e_{i}){\rm E}h'(PA_d({\rm X}_i))|\\
&=\sum_{h,h'}|{\rm E}h(PA({\rm X}_i))h'(PA_d({\rm X}_i))\\
&- {\rm E}h(PA({\rm X}_i)){\rm E}h'(PA_d({\rm X}_i))|.
\end{aligned}
\end{equation}

There are two cases: (i)$PA_d({\rm X}_i) \equiv PA({\rm X}_i)$ (ii) $PA_d({\rm X}_i) \subseteq PA({\rm X}_i)$. In case (i), the random noise $e_{i}$ and $h'(PA_d({\rm X}_i))$ are independent, the value of $dep({\rm X_i},PA_d({\rm X}_i))$ is equal to $\sum_{h'}{\rm Var}(h'(PA({\rm X}_i)))$. In case (ii), the set of determined parents of ${\rm X}_i$ is a subset of parents of ${\rm X}_i$. Then the value of $|{\rm E}h(PA({\rm X}_i))h'(PA_d({\rm X}_i))- {\rm E}h(PA({\rm X}_i)){\rm E}h'(PA_d({\rm X}_i))|$ is smaller than the square root of ${\rm Var}(h(PA({\rm X}_i))){\rm Var}(h'(PA_d({\rm X}_i)))$ based on the covariance inequality. According to the condition variance identity\citep{casella2021}, we can unfold ${\rm Var}(h'(PA_d({\rm X}_i)))$ to ${\rm E}({\rm Var}(h'(PA_d({\rm X}_i))|PA({\rm X}_i)))+{\rm Var}({\rm E}(h'(PA_d({\rm X}_i))|PA({\rm X}_i)))$. Given $PA({\rm X}_i)$, the expectation of condition variance ${\rm Var}(h'(PA_d({\rm X}_i))|PA({\rm X}_i))$ is equal to ${\rm E}(h'(PA_d({\rm X}_i))-{\rm E}(h'(PA_d({\rm X}_i))|PA({\rm X}_i)))^2=0$ due to $h'(PA_d({\rm X}_i))$ can be represented by $h(PA({\rm X}_i))$ and the variance of condition expectation ${\rm E}(h'(PA_d({\rm X}_i))|PA({\rm X}_i)))$ is equal to ${\rm Var}(h(PA({\rm X}_i)))$. Hence, in case (ii), the value of $dep({\rm X_i},PA_d({\rm X}_i)))$ is smaller than $\sum_{h'}{\rm Var}(h'(PA({\rm X}_i))$.

Step (2): For variable ${\rm X}_i$, assume that $d-1$ variables are identified as the parents of ${\rm X}_i$, which have been added to the set of determined parents of ${\rm X}_i$.

Step (3): Consider the different order dependence level of the $d^{th}$ variable ${\rm X}_d$ and ${\rm X}_i$ based on the determined parents $PA_d({\rm X}_i)$. If the value of $dep({\rm X}_i,{\rm X}_d)$ is equal to zero, then ${\rm X}_d$ and ${\rm X}_i$ are independent. Otherwise, if ${\rm X}_d$ is one of the parents of ${\rm X}_i$, then $dep({\rm X}_i,({\rm X}_d,PA_d({\rm X}_i)))=\sum_{h'}{\rm Var}(h'({\rm X}_d,PA_d({\rm X}_i)))=\sum_{h'}{\rm Var}h'(PA({\rm X}_i))\geq dep({\rm X}_i,PA_d({\rm X}_i))$ , which have been proved in Step (1). If variable ${\rm X}_d$ is not one of the parents of ${\rm X}_i$, then one has $PA_d({\rm X}_i) \equiv PA({\rm X}_i)$ and $dep({\rm X}_i,({\rm X}_d,PA_d({\rm X}_i)))\leq \sum_{h,h'}({\rm Var}(h(PA({\rm X}_i))){\rm Var}(h'({\rm X}_d,PA_d({\rm X}_i))))^{\frac{1}{2}}$ based on the covariance inequality. Considering that the value of $dep({\rm X}_i,{\rm X}_d)$ is not equal to zero, variable ${\rm X}_d$ can be represented by ${\rm X}_i$. Thus, we can unfold ${\rm Var}(h'({\rm X}_d,PA_d({\rm X}_i)))$ to ${\rm E}({\rm Var}(h'({\rm X}_d,PA_d({\rm X}_i))|PA({\rm X}_i)))+{\rm Var}({\rm E}(h'({\rm X}_d,PA_d({\rm X}_i))|PA({\rm X}_i)))$. In the same manner used in Step (1), one has that ${\rm Var}(h'({\rm X}_d,PA_d({\rm X}_i)))$ is equal to ${\rm Var}h'(PA({\rm X}_i))$ and $dep({\rm X}_i,({\rm X}_d,PA_d({\rm X}_i))) \leq \sum_{h'}{\rm Var}h'(PA({\rm X}_i)) = dep({\rm X}_i,PA_d({\rm X}_i))$.

Hence, we can correctly identify parents for each variable by comparing the different order of dependence level between $dep({\rm X}_i,({\rm X}_d,PA_d({\rm X}_i)))$ and $dep({\rm X}_i,PA_d({\rm X}_i))$, where the value of $dep({\rm X}_i,{\rm X}_d)$ is not zero and ${\rm X}_d\nsubseteq PA_d({\rm X}_i)$. By mathematical induction, this completes the proof.$\square$
\subsection*{A.2 Proof of Proposition 1}
\noindent
{\bf Proof}. Let $\left \{\phi _r\right \}_{r=1}^{n}$ be an orthonormal basis of $\mathcal{H}_{{\rm X}_i}$. Then any $f_i\in \mathcal{H}_{{\rm X}_i}$ can be written uniquely
\begin{equation}
f_i(\cdot)=\sum_{r=1}^{n}\alpha_r\phi_r(u),\ \alpha_r\in R\label{eq:8}
\end{equation}

It is indicated that the difference on $f\in \mathcal{H}_{{\rm X}_i}$ depends on $\alpha_r$. According to the definition of cross-covariance $\left\langle f_i,C_{{\rm X}_i{\rm X}_j}f_j\right\rangle$, the process of calculation only focus on the feature map of two separable reproducing kernel Hilbert space, regardless of specificity of $f_i$ and $f_j$ on reproducing kernel Hilbert spaces. Then the relative change of $\text{HSIC}(f_i({\rm X}_i),f_j({\rm X}_j))$ can be approximated by $\text{HSIC}({\rm X}_i,{\rm X}_j)$ as specified $f_i({\rm X}_i)={\rm X}_i$ and $f_i({\rm X}_j)={\rm X}_j$.

\subsection*{A.3 Proof of Proposition 2}
\noindent
{\bf Proof}. \sloppy As is defined in Definition 6, we can unfold $\mbox{HSIC}({\rm X}_i,{\rm\bf X})$ like following:
\begin{equation}
\begin{aligned}
&\mbox{HSIC}({\rm X}_i,{\rm \bf X})=\\
&\left\|C_{{\rm X}_i{\rm \bf X}}\right\|_{HS}^{2}=\left\|{\rm\bf E}_{{\rm X}_i{\rm \bf X}}[(\phi({\rm X}_i)-\mu_{{\rm X}_i})\otimes (\psi({\rm \bf X})-\mu_{{\rm \bf X}})]\right\|_{HS}^{2}\\
\end{aligned}
\end{equation}
where $\mu_{{\rm \bf X}}:=\rm\bf{E}_{{\rm \bf X}}\psi ({\rm \bf X})$. We can approximate $\psi ({\rm \bf X})$ as $\psi(a)+\sum_{j=1}^{n}\psi'_{{\rm X}_j}(a)({\rm X}_j-a)+\frac{1}{2}\sum_{j=1}^{n}\psi''_{{\rm X}_j{\rm X}_j}(a)({\rm X}_{j}-a)^{2}+2\sum_{j\neq m}^{n}\psi''_{{\rm X}_j{\rm X}_m}(a)({\rm X}_j-a)({\rm X}_m-a)$ by second-order Taylor polynomial of a function of multivariables. On the other hand, we cab approximate $\text{HSIC}({\rm X}_i,\sum_{j=1}^{n}{\rm X}_j)$ like following:
\begin{equation}
\begin{aligned}
&\mbox{HSIC}({\rm X}_i,\sum_{j=1}^{n}{\rm X}_j)\\
&=\left\|C_{{\rm X}_i\sum_{j=1}^{n}{\rm X}_j}\right\|_{HS}^{2}=\\
&\left\|{\rm\bf E}_{{\rm X}_i(\sum_{j=1}^{n}{\rm X}_j)}[(\phi({\rm X}_i)-\mu_{{\rm X}_i})\otimes (\psi(\sum_{j=1}^{n}{\rm X}_j)-\mu_{\sum_{j=1}^{n}{\rm X}_j})]\right\|_{HS}^{2}\\
\end{aligned}
\end{equation}
where $\mu_{\sum_{j=1}^{n}{\rm X}_j}:=\rm\bf{E}_{\sum_{j=1}^{n}{\rm X}_j}\psi (\sum_{j=1}^{n}{\rm X}_j)$. We can approximate $\psi (\sum_{j=1}^{n}{\rm X}_j)$ as $\psi(b)+\psi'(b)(\sum_{j=1}^{n}{\rm X}_j-b)+\frac{1}{2}\psi''(b)((\sum_{j=1}^{n}{\rm X}_j)-b)^2 \approx \psi(b)+\sum_{j=1}^{n}\psi'_{{\rm X}_j}(b)({\rm X}_j-b/n)+\frac{1}{2}\sum_{j=1}^{n}\psi''_{{\rm X}_j{\rm X}_j}(b)({\rm X}_{j}-b/n)^{2}+\sum_{j\neq m}^{n}\psi''_{{\rm X}_j{\rm X}_m}(b)({\rm X}_j-b/n)({\rm X}_m-b/n)$ by second-order Taylor polynomial of a function of single variable. Hence, we can infer that $\mbox{HSIC}({\rm X}_i,{\rm\bf X})\approx \text{HSIC}({\rm X}_i,\sum_{j=1}^{n}{\rm X}_j)$ for ${\rm\bf X}=\{{\rm X}_1,...,{\rm X}_n\}$.

\subsection*{A.4 Proof of Proposition 3}
\noindent
{\bf Proof}. We first assume that ${\rm X}_j \in PA({\rm X}_i)$, that is ${\rm X}_i= f_{i}(PA({\rm X}_i))+e_i$ for some $f_{i} \in \mathcal{H}_{{\rm X}_t}$. On the assumption that ${\rm X}_i$ is obtained by ANMs, we have $\text{HSIC}({\rm X}_j,{\rm X}_i) = \text{HSIC}({\rm X}_j, f_i(PA({\rm X}_i))+e_i)$. For a further simplification on the formulation based on Proposition 1 and 2, $\text{HSIC}({\rm X}_j, f_i(PA({\rm X}_i))+e_i)$ can be approximated by $\text{HSIC}({\rm X}_j,\sum_{{\rm X}_t\in PA({\rm X}_i)}{\rm X}_t)$. Then we unfold the formula as the following according to additive cross-covariance as is shown in Eq.\ref{eq:5}:
\begin{equation}
\begin{aligned}
\text{HSIC}({\rm X}_j,\sum_{{\rm X}_t\in PA({\rm X}_i)}{\rm X}_t)&=\left\|C_{{\rm X}_j\sum_{{\rm X}_t\in PA({\rm X}_i)}{\rm X}_t}\right\|_{HS}^2 \\
&=\left\|\sum_{{\rm X}_t\in PA({\rm X}_i)} C_{{\rm X}_j{\rm X}_t}\right\|_{HS}^{2}\label{eq:9}
\end{aligned}
\end{equation}
Then we can unfold $\text{HSIC}({\rm X}_k,{\rm X}_i)$ as the same manner
\begin{equation}
\text{HSIC}({\rm X}_k,\sum_{{\rm X}_t\in PA({\rm X}_i)}{\rm X}_t)=\left\|\sum_{{\rm X}_t\in PA({\rm X}_i)}C_{{\rm X}_k{\rm X}_t}\right\|_{HS}^{2}\label{eq:10}
\end{equation}
Since ${\rm X}_j\in PA({\rm X}_i)$, then there must be one term on Eq.\ref{eq:9} for calculating $C_{{\rm X}_j{\rm X}_j}$, which is much larger than other terms $C_{{\rm X}_j{\rm X}_t}$ for $t \neq j$ in Eq.\ref{eq:9} and $C_{{\rm X}_k{\rm X}_t}$ in Eq.\ref{eq:10}. Thus,  $\text{HSIC}({\rm X}_j, {\rm X}_i)>\text{HSIC}({\rm X}_k,{\rm X}_i)$ for ${\rm X}_j \in PA({\rm X}_i)$, ${\rm X}_k \notin PA({\rm X}_i)$ and ${\rm X}_i \notin PA({\rm X}_k)$.

For the case of ${\rm X}_i \in PA({\rm X}_j)$, the proof is same as the case of ${\rm X}_j \in PA({\rm X}_i)$.

\subsection*{A.5 Proof of Proposition 4}
\noindent
{\bf Proof}.
According to Proposition 1 and 2, we can use $\text{HSIC}({\rm X}_j+{\rm X}_k,{\rm X}_i)$ to approximate second-order $\text{HSIC}(({\rm X}_j,{\rm X}_k),{\rm X}_i)$.\\

Then, for part (a), based on the Proposition 3 and bilinear property of inner product, we first unfold $\text{HSIC}({\rm X}_j+{\rm X}_k,{\rm X}_i)$ as:
\begin{equation}
\begin{aligned}
&\text{HSIC}({\rm X}_j+{\rm X}_k,{\rm X}_i)\\
&=\left\|\sum_{{\rm X}_t\in PA({\rm X}_i)}C_{{\rm X}_j{\rm X}_t}+\sum_{{\rm X}_t\in PA({\rm X}_i)}C_{{\rm X}_k{\rm X}_t}\right\|_{HS}^{2}\label{eq:12}
\end{aligned}
\end{equation}
Since Hilbert-Schmidt norm is the norm for Hilbert space\cite[theorem~ 12.1.1]{Aubin2011}, then following the property of norm space we have
\begin{equation}
\begin{aligned}
&\left\|\sum_{{\rm X}_t\in PA({\rm X}_i)}C_{{\rm X}_j{\rm X}_t}+\sum_{{\rm X}_t\in PA({\rm X}_i)}C_{{\rm X}_k{\rm X}_t}\right\|_{HS}^{2}\leq \\
&(\left\|\sum_{{\rm X}_t\in PA({\rm X}_i)}C_{{\rm X}_k{\rm X}_t}\right\|_{HS}+\left\|\sum_{{\rm X}_t\in PA({\rm X}_i)}C_{{\rm X}_j{\rm X}_t}\right\|_{HS})^2
\end{aligned}
\end{equation}
Since ${\rm X}_k\notin PA({\rm X}_i)$, then the term $\left\|\sum_{{\rm X}_t\in PA({\rm X}_i)}C_{{\rm X}_k{\rm X}_t}\right\|_{HS}$ in the right-hand side of the inequality is tend to be zero, that is
\begin{equation}
\begin{aligned}
&\left\|\sum_{{\rm X}_t\in PA({\rm X}_i)}C_{{\rm X}_j{\rm X}_t}+\sum_{{\rm X}_t\in PA({\rm X}_i)}C_{{\rm X}_k{\rm X}_t}\right\|_{HS}^{2}\leq\\
& \left\|\sum_{{\rm X}_t\in PA({\rm X}_i)}C_{{\rm X}_j{\rm X}_t}\right\|_{HS}^2
\end{aligned}
\end{equation}
For part (b), based on the Proposition 3 and bilinear property of inner product,, $\text{HSIC}({\rm X}_j+{\rm X}_s, {\rm X}_i)$ can be unfold as
\begin{equation}
\begin{aligned}
&\text{HSIC}({\rm X}_j+{\rm X}_s,{\rm X}_i)=\\
&\left\|\sum_{{\rm X}_t\in PA({\rm X}_i)}C_{{\rm X}_j{\rm X}_t}+\sum_{{\rm X}_t\in PA({\rm X}_i)}C_{{\rm X}_s{\rm X}_t}\right\|_{HS}^{2}\label{eq:11}
\end{aligned}
\end{equation}
which is larger than $max\left\{ \text{HSIC}({\rm X}_j,{\rm X}_i),\text{HSIC}({\rm X}_s,{\rm X}_i)\right\}$ for ${\rm X}_j, {\rm X}_s \in Pa({\rm X}_i)$

For part (c), it can infer based on Proposition 3, (a) and (b).

\bibliographystyle{plainnat}
\bibliography{references}

\begin{IEEEbiographynophoto}{Yafei WANG}
is a PhD student at Shanghai University of Finance and Economics in P.R.China, supervised by Prof. Jianguo LIU. She currently works on understanding and improving Bayesian network, as well as applying Bayesian network to predict the sytemic risk in financial networks. 
\end{IEEEbiographynophoto}
\begin{IEEEbiographynophoto}{Jianguo LIU}
is a Proffessor of School of Accountancy at Shanghai University of Finance and Economics in P.R.China. His research interests include Business data analytics, Knowledge management, Social Big Data.
\end{IEEEbiographynophoto}

\vfill

\end{document}